%% file: egpaper_for_review.tex
\documentclass[10pt,twocolumn,letterpaper]{article}

\usepackage{cvpr} 
\usepackage{times}
\usepackage{epsfig}
\usepackage{graphicx}
\graphicspath{ {./figures/} }
\usepackage{amsmath}
\usepackage{amssymb}
\usepackage{makecell} 
\usepackage{booktabs}
\usepackage{multirow}
\usepackage{hyperref}
\usepackage{float}
\usepackage{cleveref}

\begin{document}

\title{GaitFace: A Multimodal Dataset for Long-Range Person Identification}

\author{
Alain Komaty \quad
Luis S. Luevano \quad
Vidit Vidit \quad
Zeina Al Amine \quad
Anjith George \quad
S\'ebastien Marcel \\
Idiap Research Institute, Martigny, Switzerland \\
{\tt\small \{akomaty, luis.luevano, vidit.vidit, anjith.george, zeina.alamine, sebastien.marcel\}@idiap.ch}
}

\maketitle
\thispagestyle{empty}

\begin{abstract}
Efficient border control is becoming a significant global challenge, mainly due to severe congestion and extended passenger waiting times. To mitigate these bottlenecks and facilitate passenger flow, biometric technologies are increasingly deployed to streamline identity verification and enhance crossing efficiency. Technical limitations frequently impede biometric identification, particularly in long-range surveillance, where systems must deal with adverse atmospheric conditions and degraded image quality. While high-quality frameworks like BRIAR exist, they are frequently restricted to specific government agencies. This paper introduces GaitFace, a new public dataset that contains face and gait data captured at long distances. To ensure that the research reflects authentic border scenarios, we use Pre-Enrollment data, where a traveler registers via a mobile device, and "In-the-Wild" captures, which records individuals at a distance across multiple viewing angles and different cameras. Benchmarking SOTA face and gait models reveals that current architectures fail under low-resolution and elevated viewpoints despite success with optical assistance. GaitFace exposes these critical vulnerabilities, providing a rigorous public benchmark to drive more robust, unconstrained biometric research. Available at \url{https://idiap.ch/paper/gaitface} 
\end{abstract}

\vspace{-1.5em}
\section{Introduction}
Biometric recognition at long distances is an essential capability for automated border crossing \cite{border_control_survey_labati_2016, border_crossing_cbp_khan_2021}, surveillance \cite{lr_surveillance_martinez-diaz_2021}, and forensic \cite{forensic_gait_seckiner_2019} applications. Unlike traditional close-range biometrics in controlled scenarios, long-distance person recognition presents scenarios with severe image degradation caused by low resolution, motion blur, changing viewpoints, atmospheric effects, and varying illumination and weather conditions.

Face and gait data are particularly relevant in this setting because both can be used to provide automatic biometric recognition. These two modalities also provide complementary information: face recognition is highly discriminative when sufficient facial detail is available, while gait can remain informative at long distances.

\input{figures/gait_wtc_example}



Despite recent progress in face and gait recognition, existing datasets only partially cover the conditions required for realistic long-distance person recognition. Large-scale face datasets collected in the wild have enabled the development of highly-accurate face recognition models\cite{martinez-diaz_benchmarking_fr_2021}, but they are driven by high-quality face imagery and do not show the same performance with native low-quality surveillance probes\cite{lr-survey-luevano-2019}. Low quality and surveillance face datasets provide more challenging probes, but often lack controlled high-quality enrollment data \cite{tinyface, survface}, long-distance optical-zoom captures\cite{scface}, or temporal variability across repeated sessions\cite{ijbs}. Similarly, most gait datasets~\cite{casiab,gait3d,ccpg,guo2025gait} include controlled or short-to-medium range sequences, while long-distance gait probes with realistic viewpoint, clothing, and accessory changes remain scarce. As a result, current benchmarks encounter limitations in designing person recognition protocols. Another important constraint is dataset availability. Some of the most relevant resources for long-range and surveillance biometrics, such as IJB-S\cite{ijbs} and BRIAR\cite{briar}, are highly restricted, limiting reproducibility and excluding many academic, international, and commercial users. At the same time, biometric dataset collection must address personal privacy, informed consent, and comply with data governance regulations \cite{gdpr2016, ai_act_2025}.

To address these gaps, we introduce \textbf{GaitFace}, a multimodal dataset for long-range person identification designed around a border-crossing-inspired acquisition protocol. GaitFace combines controlled pre-enrollment face captures from mobile devices with outdoor long-distance observations, including high-resolution optical-zoom face images, native low-resolution face video, and long-distance gait recordings from multiple viewpoints. The dataset contains 70 participants recorded across two sessions on different days, introducing temporal variability in clothing, accessories, hairstyle, subject appearance, and environmental conditions. This design enables the study of high-quality reference to long-distance probe matching, low-resolution face recognition, long-distance gait recognition, and multimodal face-gait analysis. The main contributions in this paper are:

\begin{itemize}
    \item We introduce \textbf{GaitFace}, a multimodal long-distance person recognition dataset combining controlled mobile pre-enrollment, high-resolution optical-zoom face captures, native low-resolution face video, and long-distance gait recordings.
    \item We design a realistic multi-session acquisition protocol with multiple viewpoints, subject motion, clothing changes, accessories, phone usage, and natural environmental variability.
    \item We define evaluation protocols for long-distance face recognition, long-distance gait recognition, and cross-session matching under realistic covariate changes.
    \item We provide benchmark results using state-of-the-art face and gait recognition methods, highlighting the difficulty of the proposed setting and establishing baselines for future research.
\end{itemize}
The paper is organized as follows: Section~2 reviews related work in long-distance face and gait recognition. Section~3 presents our proposed GaitFace dataset. Section~4 describes the experimental setup and evaluation protocols. Section~5 reports benchmark results for face and gait recognition. Finally, we give our concluding remarks in Section~6.


\section{Related Work}

In this section, we review prior work across long-distance face and gait recognition, highlighting the critical gaps in publicly available multimodal benchmarks to motivate our proposed dataset.

\textbf{Unimodal Recognition Limitations} \\
Modern face recognition relies heavily on large-scale, high-resolution datasets (e.g., LFW \cite{lfw}, CASIA-WebFace \cite{casia-webface}, MS-Celeb-1M \cite{msceleb1m}, WebFace260M \cite{zhu2021webface260m}, Glint360K \cite{partialfc-glint360k}) and benchmarks like IJB-C \cite{ijbc}, but models trained on these suffer severe performance drops when applied to low-resolution surveillance scenarios (e.g., SCFace \cite{scface}, UCCS \cite{uccs}, DroneFace \cite{droneface}, TinyFace \cite{tinyface}, SurvFace \cite{survface})
Similarly, unimodal gait recognition methods are typically evaluated on datasets restricted to short-to-medium ranges (e.g., CASIA-B \cite{casiab}, Gait3D \cite{gait3d})

\textbf{Multimodal Face andn Gait Datasets} \\
Because facial features degrade at a distance while gait remains informative, multimodal evaluation is critical for realistic border control and surveillance
However, publicly available datasets rarely provide synchronized face and gait captures at extreme distances. While datasets like IJB-S \cite{ijbs} and BRIAR \cite{briar, liu2025person, wang2025combo} capture long-range surveillance scenarios, their access is highly restricted, limiting broad research and commercial reuse. To address this gap, GaitFace introduces a public benchmark combining controlled mobile pre-enrollment with long-distance multimodal outdoor captures. Table \ref{tab:multimodal_datasets} compares GaitFace against the limited existing multimodal datasets from the literature.

\begin{table}[htbp]
\centering
\resizebox{\columnwidth}{!}{%
\begin{tabular}{l l l c l}
\hline
\textbf{Dataset Name} & \textbf{Dist} & \textbf{Env} & \textbf{\#Subjs} & \textbf{Public} \\ 
\hline
BRIAR \cite{briar, liu2025person, wang2025combo} & $<$1,000m & Both & $\sim$1,760 & No \\ 
NIST Database \cite{kale2004fusion} & 10-15m & Outdoor & 30 & Yes \\ 
OU-ISIR Database \cite{muramatsu2013multi} & $\sim$8m & Indoor & 1,912 & Yes \\ 
FOCS \cite{maity2021multimodal} &  $<$8.3m & Outdoor & 437 & Yes \\ 
NLPR Gait \cite{geng2008adaptive} & 5-12m & Outdoor & 20 & Yes \\ 
CASIA-B-Gait-Face \cite{fu2022fusion} & 5-15m & Indoor & 124 & Yes \\ 
Walking Video \cite{aung2022multimodal} & $<$25m & Outdoor & 25 & No \\ 
\textbf{GaitFace (Ours)} & \textbf{ $<$100m} & \textbf{Both} & \textbf{70} & \textbf{Yes} \\
\hline
\end{tabular}%
}
\caption{Comparison of Multimodal Face and Gait Datasets}
\label{tab:multimodal_datasets}
\end{table}

\section{GaitFace: Long-distance face \& gait dataset}
\paragraph{Overview} The dataset was ethically and legally acquired from 70 consenting participants (ID 0–69) under a protocol approved by the relevant institutional review boards, with all individuals providing explicit informed consent for the collection and processing of their biometric data. Recorded across two acquisition sessions separated by several weeks, the dataset includes associated demographic metadata and amounts to approximately 2.7 TB of multi-modal data, comprising RGB high-resolution facial images in RAW format and multi-view gait recordings. The collection protocol incorporates repeated measurements to capture natural temporal variations in appearance and utilizes a long-distance acquisition design—capturing biometrics at distances of up to 100 meters—to reflect realistic border surveillance conditions. By recording face and gait data simultaneously from multiple viewpoints, the dataset enables consistent multi-modal analysis for each participant, and it will be publicly released upon acceptance.

\subsection{Capture conditions and devices}

Data were collected outdoors, ensuring realistic variability in illumination (natural daylight variations) and weather conditions (sunny, cloudy, windy, and rainy). To introduce controlled variations in appearance and behavior, participants were recorded under four standardized scenarios: normal walking, walking while simulating a phone call, walking with an accessory (hand bag or backpack), and walking with a different jacket. All recordings followed a predefined walking trajectory (Figure~\ref{subfig:map_positions}) shared across scenarios and sessions: participants started from a fixed point, walked along the path to a designated stop location where they briefly paused and faced the third floor camera (Figure~\ref{subfig:camera_resolution}), continued to an end point, turned, and returned to the starting point. This round-trip path was repeated for each scenario. The acquisition geometry includes a ground-level viewpoint (eye-level perspective) and an elevated viewpoint (approximately 9--10 meters), simulating typical surveillance camera positions.


\subsection{Data Capture Modalities}
\begin{figure}[h]
    \centering
    \setlength{\tabcolsep}{1pt}

    \begin{subfigure}{0.19\columnwidth}
        \centering
        \includegraphics[width=\textwidth]{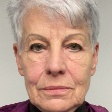}
        \caption{}
        \label{fig:enroll}
    \end{subfigure}%
    \begin{subfigure}{0.19\columnwidth}
        \centering
        \includegraphics[width=\textwidth]{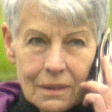}
        \caption{}
        \label{fig:probe_hq}
    \end{subfigure}%
    \begin{subfigure}{0.19\columnwidth}   
        \centering
        \includegraphics[width=\textwidth]{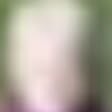}
        \caption{}
        \label{fig:probe_100_3f}
    \end{subfigure}%
    \begin{subfigure}{0.19\columnwidth}
        \centering
        \includegraphics[width=\textwidth]{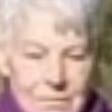}
        \caption{}
        \label{fig:probe_100_gf}
    \end{subfigure}%
    \begin{subfigure}{0.19\columnwidth}
        \centering
        \includegraphics[width=\textwidth]{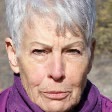}
        \caption{}
        \label{fig:probe_3m_gf}
    \end{subfigure}

    \caption{Data samples: (a) Enroll, (b) Probe HQ, (c) 100m $3^{rd}$ Floor (3F), (d) 100m Ground Floor (GF), (e) 3m GF.}
    \label{fig:face_samples_enroll_probe}
\end{figure}

We describe the multiple modalities considered in GaitFace which are: pre-enrollment face captures, outdoor face captures, outdoor gait captures, and relevant subject metadata.
\vspace{-1em}
\paragraph{Pre-enrollment Face Captures.} Face enrollment (Figure~\ref{fig:face_samples_enroll_probe}a) was performed indoors using two mobile devices (iPhone 12 and Samsung Galaxy S9), where each participant recorded facial sequences for 15 seconds, including static (first 5 seconds of the recording) and controlled horizontal and vertical head movements (last 10 seconds of the recording) using the phones' frontal camera. These data serve as high-quality ground-truth references for subsequent analysis. This setup mimics a real-world remote pre-enrollment process.


\vspace{-2em}
\paragraph{Outdoor Face Captures}
 Facial data were collected in an outdoor environment to simulate realistic long-distance surveillance conditions under natural illumination. Participants followed the predefined path and were instructed to stop at a designated point and face the sensor for a short duration (5--10 seconds), enabling the acquisition of high-quality facial images using a long-range zoom camera (Figure~\ref{subfig:camera_resolution}) positioned at the third-floor level (Figure~\ref{fig:face_samples_enroll_probe}b). The captured images were stored in RAW format to preserve fine-grained facial details. Simultaneously, the faces were recorded from two cameras (Canon EOS 90D and Fujifilm X-T4 shown in Figure~\ref{subfig:camera_resolution}) from the $3^{rd}$ floor (Figure~\ref{fig:face_samples_enroll_probe} (c)) and Ground floor (Figure~\ref{fig:face_samples_enroll_probe} (d),(e)). The acquisition was conducted under varying weather conditions, including sunny, cloudy, and rainy environments, introducing natural variability in lighting and visibility. This setup provides high-quality reference imagery, enabling robust evaluation of face recognition systems under challenging conditions.
 
\begin{figure}[h]
\centering
\setlength{\tabcolsep}{1pt}

\begin{tabular}{cccc}
\includegraphics[page=1,width=0.11\textwidth]{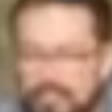} &
\includegraphics[page=1,width=0.11\textwidth]{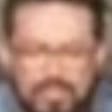} &
\includegraphics[page=1,width=0.11\textwidth]{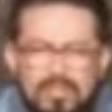} &
\includegraphics[page=1,width=0.11\textwidth]{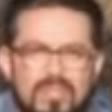} \\

\includegraphics[page=1,width=0.11\textwidth]{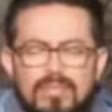} &
\includegraphics[page=1,width=0.11\textwidth]{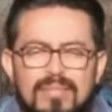} &
\includegraphics[page=1,width=0.11\textwidth]{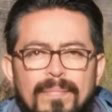} &
\includegraphics[page=1,width=0.11\textwidth]{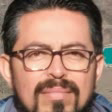}
\end{tabular}

\caption{Face captures acquired using the GF camera, showing the subject's face evolution while approaching the camera at different timestamps (Top left: first frame, bottom right: last frame).}
\label{fig:scenario2_frames}
\end{figure}

\noindent\textbf{Outdoor Gait Captures}
Gait data were collected in an outdoor environment using a multi-view acquisition setup designed to capture full-body motion from complementary perspectives. Participants followed the predefined walking trajectory, ensuring consistent spatial alignment across recordings. Data were acquired simultaneously from ground-level cameras and an elevated viewpoint (third-floor level), providing both frontal/lateral and top-down observations of motion patterns. The recordings from the third-floor level were consistently captured across both sessions using the same cameras setup. At ground level, gait videos were recorded using a Fujifilm X-T4 camera, with the camera position intentionally varied between the two sessions to introduce additional viewpoint variation. Examples of the scenario of session 2 \textit{Walking Towards the Camera} (WTC) are given in Figure~\ref{subfig:gait_sequence} for gait and Figure~\ref{fig:scenario2_frames} for face. The recordings were performed under natural illumination and varying environmental conditions, introducing realistic variability in appearance and visibility. This multi-view configuration allows us to capture different gait dynamics under unconstrained outdoor settings.\\

\noindent\textbf{Quantitative Resolution Analysis.} 
To quantitatively characterize the dataset's image degradation, we measured the average face crop resolution and Inter-Pupillary Distance (IPD) across configurations. At the maximum standoff distance of 100m without optical assistance (LQ probes from GF and 3F cameras), the average face resolution is $18\times22$ pixels with an IPD of $4\pm1.2$ pixels, representing severe spatial degradation. Conversely, the 100m optically-zoomed HQ probes provide an average face resolution of $210\times250$ pixels with an IPD of $55\pm3.5$ pixels. For the WTC protocol, the face resolution scales dynamically from the baseline $18\times22$ pixels at 100m up to $450\times520$ pixels (IPD of $115\pm6.2$ pixels) at the 3m terminal range. 

\input{tables/fr_protocols}

\subsection{Metadata and demographic distribution}
\begin{figure}[t] 
    \centering
    \includegraphics[width=\linewidth]{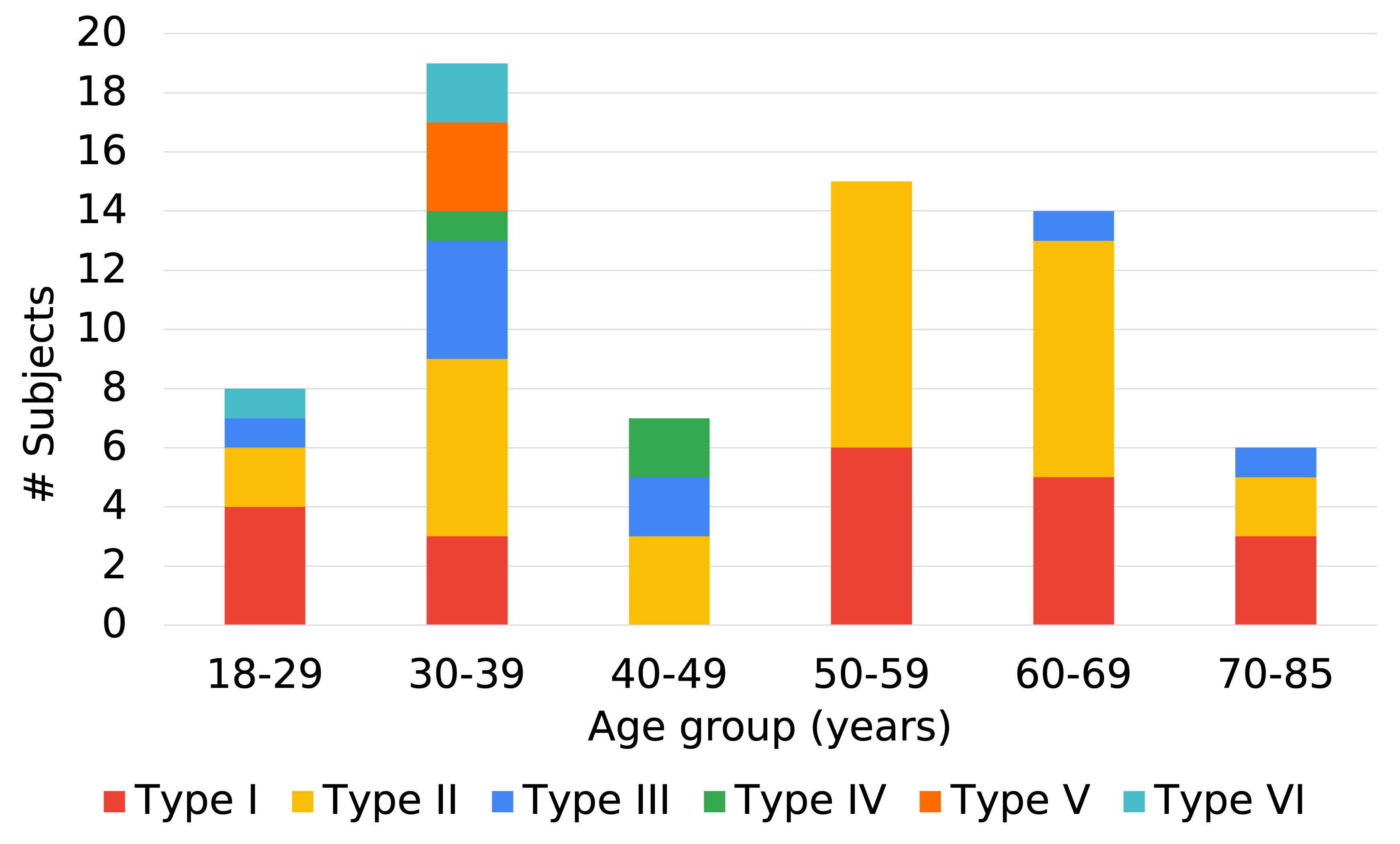}
    \caption{Demographic distributions of the dataset for Age and skin color (Fitzpatrick scale~\cite{fitzpatrik_skin_gupta_2019})}
    \label{fig:demographic_age_distribution}
\end{figure}

Along with visual data, this dataset includes participant metadata (age, gender, and skin color) to enable future benchmarking for algorithmic fairness and demographic robustness. The demographic distribution of the dataset is balanced in terms of age and gender, with 45.7\% of female and 54.3\% of male volunteers. However, due to local population characteristics at the acquisition location, the dataset predominantly represents individuals with lighter skin tones, as illustrated in Figure~\ref{fig:demographic_age_distribution}, which present the distributions of age, gender, skin color, and Fitzpatrick scale~\cite{fitzpatrik_skin_gupta_2019}.

\section{Experimental Setup}



\subsection{Face Recognition}

\paragraph{ISO/IEC 19795 Standardized Metrics}
To ensure scientific rigor and operational relevance, the performance of the LDFR system is characterized using standardized biometric metrics. These metrics allow for a precise analysis of the trade-off between security (False Match Rate (FMR)) and convenience (False Non-Match Rate (FNMR)).

\paragraph{Protocol definition}
Focusing on border scenarios, we enroll using mobile phone data (Fig.~\ref{fig:face_samples_enroll_probe}a). Probes are categorized as High-Quality (HQ, 100m with optical zoom) or Low-Quality (LQ, default lens, no optical assistance). LQ probes vary by camera position (Ground-Floor [GF] vs. 3rd Floor [3F]) and subject movement (Walking Towards Camera [WTC] from 100m to 3m). Table~\ref{tab:fr_protocols} details the specific enrollment and probe configurations for each protocol.
\input{figures/gait_views_variations}

\subsection{Gait Recognition}
The challenge in long distance gait recognition is that the motion of different parts get less detailed. Therefore, with our dataset we want to study the efficacy of gait recognition models under these difficult scenarios. Similar to other datasets~\cite{casiab,guo2025gait,ccpg},  we record sequences with variations like change of piece of clothing, carrying a bag, and gesture variation, i.e. normal walk vs walking while talking on the phone. There are two separate sessions to further include diversity in the scene and person's appearance.

\vspace{-1em}
\paragraph{Protocol and Metrics}
The walking pattern is broken down into 4 directions(\textbf{V1-V4} in \cref{fig:combined_gait_data}) where the camera sees mainly the frontal or back part of the person or the two sides. This setup is helpful to analyze models behavior when the probe and gallery sequences are from different views.
As shown in \cref{subfig:map_positions}, we record the same walking pattern from two camera levels, the third floor of a building(\textbf{G1}) and the ground floor(\textbf{G2}). The two ground floor sessions are recorded from a different location. This leads to a drastic change of viewpoint between sessions. 
The gallery consists of only one session of the normal walking sequence per id. This choice reflects authentic operational constraints: because high-quality personal gait enrollment is practically unfeasible at remote registration, the system must utilize standard baseline surveillance captures as the gallery template. The probe set is entirely made of the second session. The probe includes all the other variations that include per id (a) normal walking sequence from the second session.(b) walking with a phone. (c) carrying a bag.(d) change of clothing. The probe set is entirely made of the second session. For each sequence, we do a cross evaluation of the walking direction, for e.g. V1 direction is only compared with V2-V4 directions. 
Rank-1 accuracy is used to evaluate performance for a closed id set.

\section{Benchmarking Results}


\subsection{Long-Distance Face Recognition}
To evaluate robustness against Long Distance Face Recognition(LDFR)-induced degradation, we benchmark four SOTA models (Table~\ref{tab:fr_sota_models}) representing a computational spectrum: EdgeFace \cite{george2024edgeface} (18.22M parameters) trained on WebFace12M, for edge-deployment, AdaFace IR50/IR101 (43--65M) \cite{kim2022adaface} trained on WebFace4M and WebFace12M respectively, for quality-adaptive robustness, and LV-Face \cite{you2025lvface} (140M+) trained on Glint360K, for maximum accuracy. This selection enables a comparative analysis of how diverse architectures, ranging from lightweight hybrids to large-scale Vision Transformers, utilize specialized loss functions to mitigate the effects of atmospheric turbulence and resolution loss inherent in long-distance sensing.

\input{tables/fr_sota_models}

\paragraph{\textbf{The HQ faces using optical zoom (HQ-1, HQ-3, HQ-10 and HQ-20):}} All these protocols use the same set of probes, that is the HQ faces captured using an optical zoom attached to the camera. Table~\ref{tab:fr_HQ_results} shows the FNMR performance metrics for the four FR systems. The operational thresholds used were calibrated on IJB-C at FMR = 1\% and 0.1\%. We can notice that all the four system acheive very good recognition results, with slight advantage for LVFace and IR101. This indicates that when baseline optical resolution is maintained by the camera lens, larger models effectively leverage the preserved fine-grained facial features. Performance improves slightly as enrollment samples increase; LVFace’s FNMR drops from 0.9\% (HQ-1) to 0.4\% (HQ-20). Notably, using only three samples (HQ-3) yields results comparable to using twenty (HQ-20), suggesting a performance cap.

\input{tables/sota_fr_results}

\paragraph{\textbf{Face captured from the Ground Floor (GF) at 100m distance (GF-3, GF-10):}} In this protocol, we extracted probe frames from a video sequence. We used 1, 10, 20 and 40 frames to investigate how increasing the number of probe images would affect the performance. The overall results on this protocol were not as good as the ones from HQ protocol due to the difference of face resolution as shown in the example in figure~\ref{fig:face_samples_enroll_probe}-d.

\begin{figure}[t!]
    \centering
    \includegraphics[width=1\linewidth]{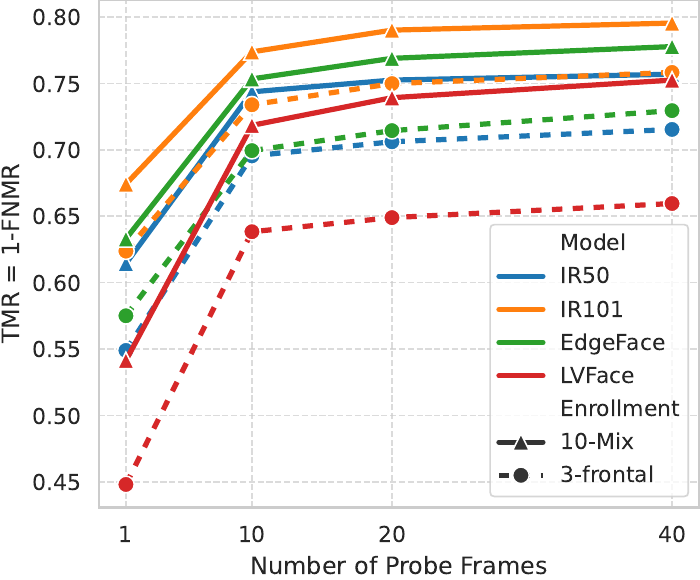}
    \caption{\textbf{Protocol GF variants (GF-3, GF-10)}: Robustness to low-quality probes through multi-frame fusion. Colors distinguish the four evaluated models (IR50, IR101, EdgeFace, LVFace). Line styles and markers denote the enrollment protocol: solid lines with triangles indicate the rich 10-Mix strategy, while dotted lines with circular markers represent the "3-frontal" strategy.}
    \label{fig:gf_trend_curves}
\end{figure}

Figure~\ref{fig:gf_trend_curves} shows the impact of probe frame count and enrollment strategy on long-distance face recognition. The plot illustrates the True Match Rate (TMR) at a pre-calibrated operational threshold (FMR=1\% on IJB-C) as the number of low-quality probe frames increases from 1 to 40. Notably, as shown in the performance trends, increasing the number of fused probe frames and utilizing a richer enrollment consistently improves recognition accuracy across all evaluated architectures. We observe that the best overall performance is achieved by the IR101 and EdgeFace models under the optimal condition of a 10-Mix enrollment combined with 40 fused probe frames. Interestingly, EdgeFace outperformed the heavier LV-Face. This suggests a critical domain gap; while SOTA models excel when provided sufficient optical detail, lightweight architectures may exhibit greater robustness against extreme resolution loss, likely avoiding the reliance on high-frequency details that hinders larger models under severe degradation.
Despite this configuration yielding the highest accuracy, the peak True Match Rate (TMR) only approximates to 80\%. 

\begin{figure}[h!]
    \centering
    \includegraphics[width=1\linewidth]{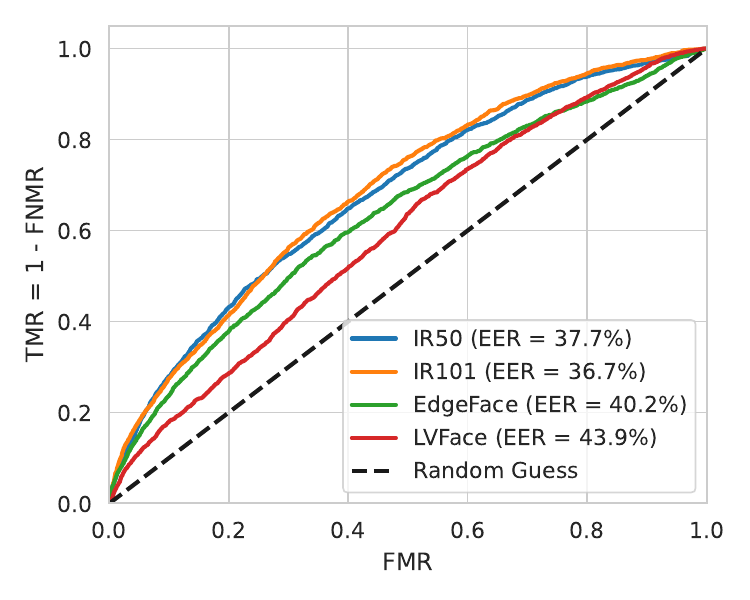}
    \caption{\textbf{Protocol 3F:} ROC curves for four face recognition systems. All curves follow the chance line (AUC $\approx 0.5$),
demonstrating failure to separate genuine and zero-effort impostor scores.}
    \label{fig:3f_roc}
\end{figure}

\vspace{-1.5em}
\paragraph{\textbf{Face captured from the 3rd Floor (3F):}} Extreme resolution loss at 100m causes severe degradation across all FR models. As shown in Fig.~\ref{fig:3f_roc}, ROC curves collapse toward the chance line, indicating a systemic inability to distinguish genuines from zero-effort impostors. This failure is reflected in high EERs: 36.7\% (IR101), 37.7\% (IR50), 40.2\% (EdgeFace), and 43.9\% (LVFace). These results demonstrate that standard architectures are practically ineffective at this range, establishing a challenging benchmark for future long-distance recognition research.

\paragraph{\textbf{Walking towards the camera (WTC):}} This protocol evaluates the effect of the distance on selected face recognition algorithm. We will use the same four recognition systems listed in table~\ref{tab:fr_HQ_results}.
For each video used in this experiment, the first frames correspond when the person was at around 100 meters away from the camera, while the last frames correspond to when the person was close to the camera (less than 4 meters) (examples in Figures~\ref{subfig:gait_sequence} and \ref{fig:scenario2_frames}).

\begin{figure}[h!]
    \centering
    \includegraphics[width=1\linewidth]{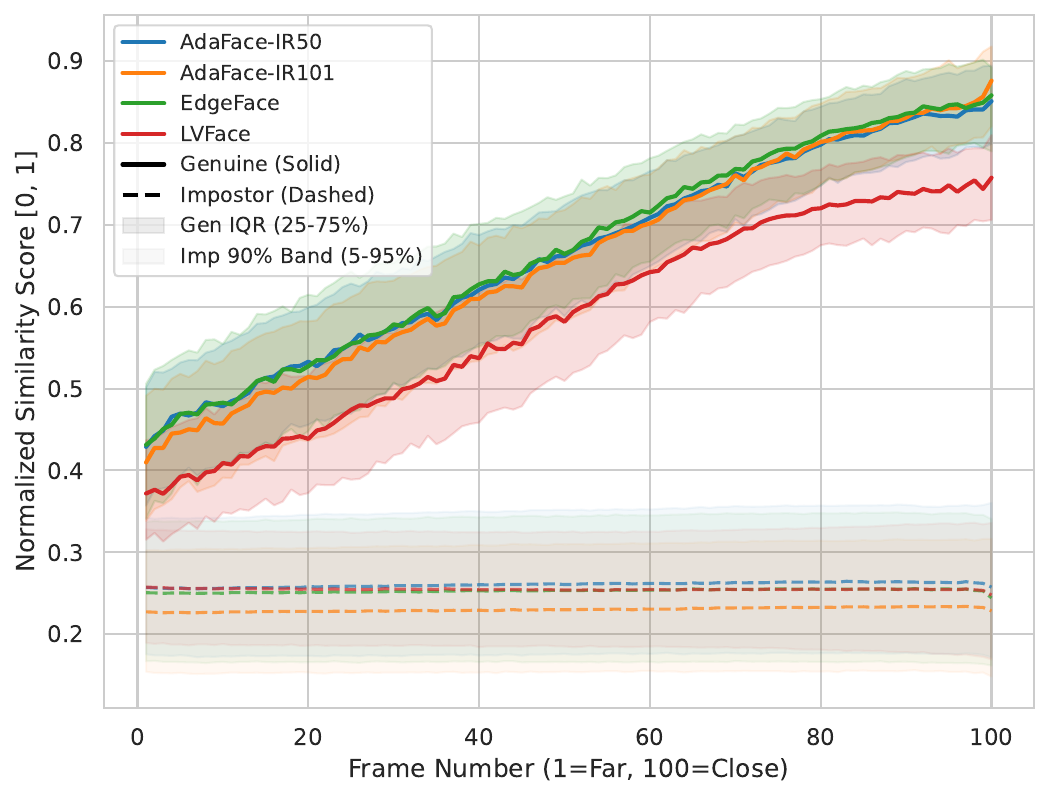}
    \caption{\textbf{Protocol WTC:} Normalized similarity score decay across distance (Frame 1 = Far, 100 = Close). Solid and dashed lines denote genuine and impostor medians, respectively, while shaded regions represent the genuine interquartile range (IQR: 25--75\%) and the impostor 90\% band (5--95\%).}
    \label{fig:perf:decay_curve}
\end{figure}

\begin{figure}[h!]
    \centering
    \includegraphics[width=1\linewidth]{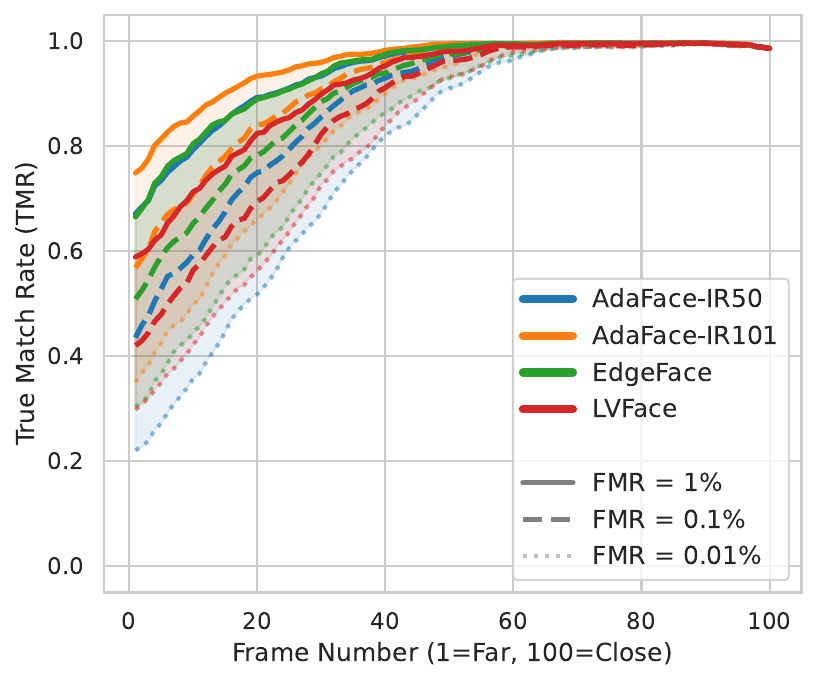}
    \caption{\textbf{Protocol WTC:} True Match Rate (TMR) degradation as a function of distance across three False Match Rate (FMR) thresholds. The shaded performance corridors bound the system accuracy from the most lenient (FMR=1\%, solid top line) to the strictest (FMR=0.01\%, dotted bottom line) operational settings. Thresholds were calibrated on IJB-C dataset.}
    \label{fig:perf:tmr_distance}
\end{figure}

Figure \ref{fig:perf:decay_curve} illustrates a monotonic increase in genuine similarity scores as subjects approach the camera (moving from Frame 1 to Frame 100). At extreme distances, the genuine interquartile ranges (IQR) and the impostor 90\% bands severely overlap, but as the subject closes the distance, these distributions successfully separate. This physical improvement in image quality and discriminability translates directly into enhanced verification accuracy, as demonstrated in Figure \ref{fig:perf:tmr_distance}. While all systems achieve near-perfect True Match Rates (TMR) at close range, their robustness degrades at farther distances and is highly sensitive to the chosen operational security level; imposing stricter thresholds (e.g., FMR = 0.01\%) triggers an earlier collapse in TMR. Notably, while the heavier architectures (AdaFace-IR101 and IR50) demonstrate the greatest resilience within this performance corridor, EdgeFace, despite being a lightweight model, consistently outperforms the heavier LVFace architecture across all distances and thresholds. Nevertheless, all evaluated models ultimately fail to sustain reliable recognition at the farthest captured frames, underscoring the profound challenge of long-distance facial recognition.

\input{figures/cmc_plots/cmc_g1}
\subsection{Long-Distance Gait Recognition}

We proceed to show baselines using state-of-the-art models on gait recognition and discuss the findings.

\input{tables/silhouette_res}

\vspace{-1.5em}
\paragraph{Baselines.} We compare the performance of  silhouette-based methods  like GaitBase~\cite{fan2023opengait},  DeepGaitV2~\cite{fan2023exploring}, and GaitGL\cite{lin2022gaitgl}. GaitBase and DeepGaitV2 both are trained on a large GREW~\cite{guo2025gait} dataset.
GaitBase~\cite{fan2023opengait} uses a ResNet-like (88M params) backbone and creates temporal and spatial features for a given sequence of silhouette frames. Although it was a simple baseline, it improved gait recognition on standard benchmarks. DeepGaitV2~\cite{fan2023exploring}(95M params) proposed novel residual layers to increase the depth of the network and improve the recognition performance.
GaitGL\cite{lin2022gaitgl}on the other hand is trained with a smaller CASIA-B\cite{casiab} dataset. 
GaitGL architecture is designed to address the limitations of traditional gait recognition by simultaneously extracting both coarse-grained global features and fine-grained local details from silhouette sequences. This model is significantly smaller number of parameters than GaitBase i.e 3.1M. Additionally, we benchmark BiggerGait~\cite{ye2025biggergait} trained on RGB-images from  CCPG~\cite{ccpg} dataset. BiggerGait uses the layer-wise features of foundation models such as  DinoV2\cite{oquab2023dinov2} to fine-tune a shallow gait encoder network. In comparison to silhouette-based methods, they achieve better performance on standard short distance gait benchmarks~\cite{ccpg,casiab} but with a parameter size of 120M.  
We use OpenGait\cite{fan2023opengait} framework to evaluate these methods on our dataset and use their publicly available trained weights. 

\vspace{-1em}
\paragraph{Results.}
We report in \cref{tab:gait_all_results}, Rank-1 accuracy across our defined protocols. Overall, both silhouette- and rgb-images-based methods generalize poorly on our dataset, highlighting the  challenges of long distance gait recognition. When the camera viewpoint between sessions is small, as in G1, the performance is higher. With respect to covariate changes such as phone, bag, or clothing, performance is worse than the normal walk sequences. However, the impact of clothing change is less in BiggerGait than in silhouette-based methods as it is trained on a large CCPG~\cite{ccpg} dataset, which contains large appearance changes. The RGB-based BiggerGait severly underperforms compared to the silhouette based method because of the domain-gap between training CCPG dataset and ours. This difference is less drastic in the binary-mask silhouettes used in GaitBase, DeepGaitV2 and GaitGL. Through, CMC plots~\cref{fig:cmc_g1,fig:cmc_g2}, we see that GaitBase~\cite{fan2023opengait} performs best under G1 protocol and DeepGaitV2~\cite{fan2023exploring}
under challenging G2 protocol. However, all of the models fail to achieve satisfactory generalization on long distance gait recognition. This encourages more research in this direction.
\section{Conclusion and Future Work}
In this paper, we introduced GaitFace, a novel multimodal dataset explicitly designed to address the  challenges of long-range person identification in realistic border crossing and surveillance scenarios. By combining controlled, high-quality mobile pre-enrollment with highly unconstrained outdoor probes, GaitFace exposes the operational vulnerabilities of current biometric systems. Our extensive benchmarking of state-of-the-art face recognition models demonstrated systemic performance degradation at extreme distances (up to 100 meters) and elevated viewpoints, despite maintaining high accuracy when assisted by optical zoom. Similarly, gait recognition baselines struggled to generalize across the viewpoint shifts, clothing variations, and carried objects introduced in our multi-session protocols.
\input{figures/cmc_plots/cmc_g2}
These baseline evaluations underscore that unconstrained long-distance biometric recognition remains an open and highly challenging problem. As future work, we plan to explore the fusion between face and gait modalities to leverage their complementary strengths, aiming to achieve reliable identification even when individual biometric cues are severely degraded by distance or atmospheric interference. We hope that the public release of the GaitFace dataset will stimulate the broader research community to develop more robust biometric solutions.

\paragraph{Acknowledgment}
The authors would like to thank all the volunteers who participated in the data collection for their continuous support for our research activities. We acknowledge the funding provided by the following projects: Frontex under the Frontex Research Grants Programme, Call 2024/CFP/INNOVATE/01, Grant Agreement No. 2025/280. CarMen project, HORIZON-CL3-2023-BM-01, no. 101168325 and PopEye project, HORIZON-CL3-2023-BM-01, no. 101168317.


{\small
\bibliographystyle{ieee}
\bibliography{egbib}
}

\end{document}

%% file: figures/gait_wtc_example.tex
\begin{figure}[!t]
    \centering
    
    \begin{subfigure}{0.95\linewidth}
        \centering
        \includegraphics[width=\linewidth]{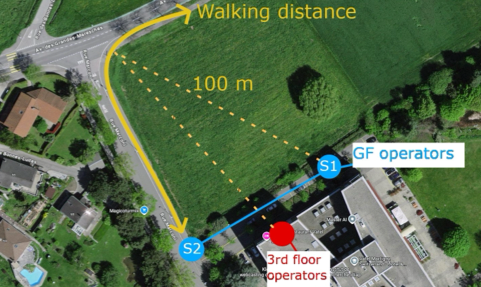}
        \caption{Map showing positions of recording teams and walking paths.}
        \label{subfig:map_positions}
    \end{subfigure}

    \begin{subfigure}{0.95\linewidth}
        \centering
        \setlength{\fboxrule}{0.5pt} 
        \setlength{\fboxsep}{0pt}
        \fbox{\includegraphics[width=\linewidth]{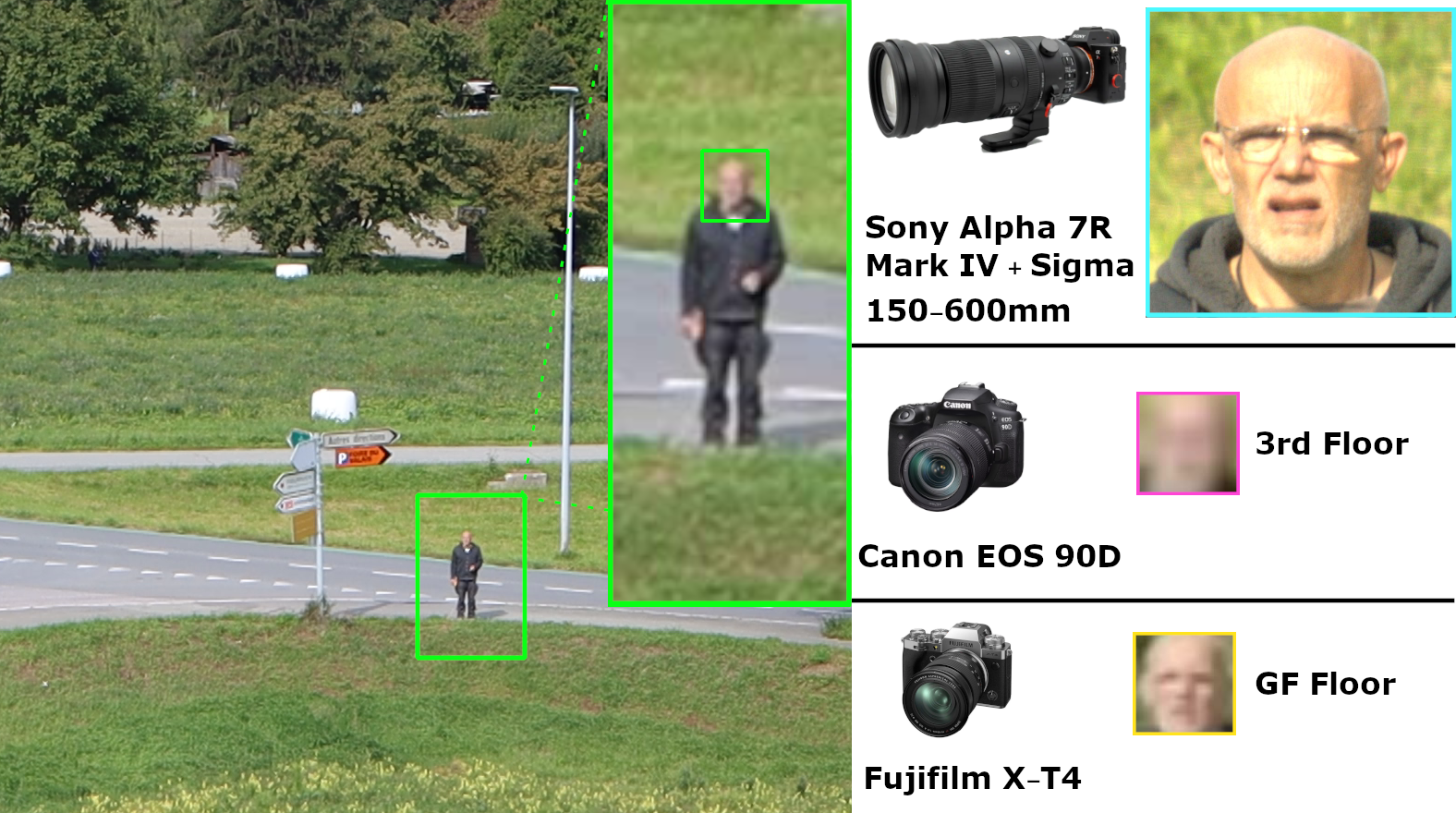}}
        \caption{Facial resolution across different cameras/heights.}
        \label{subfig:camera_resolution}
    \end{subfigure}

    \begin{subfigure}{0.95\linewidth}
        \centering
        \noindent
        \includegraphics[width=0.125\linewidth]{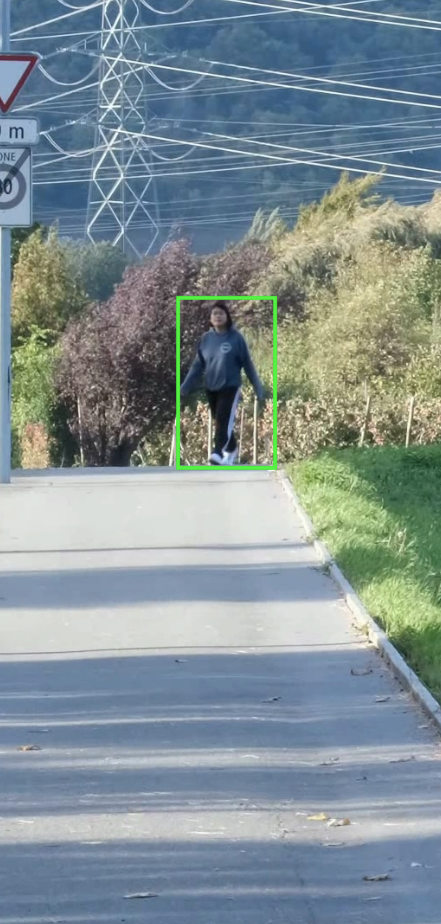}%
        \includegraphics[width=0.125\linewidth]{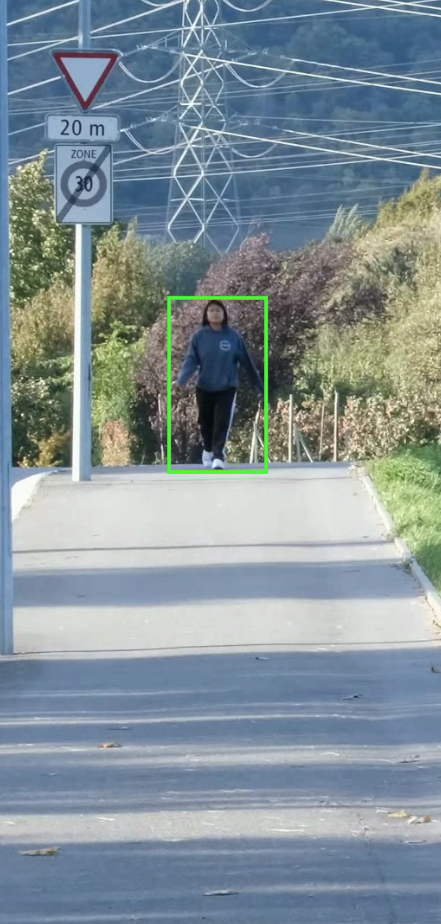}%
        \includegraphics[width=0.125\linewidth]{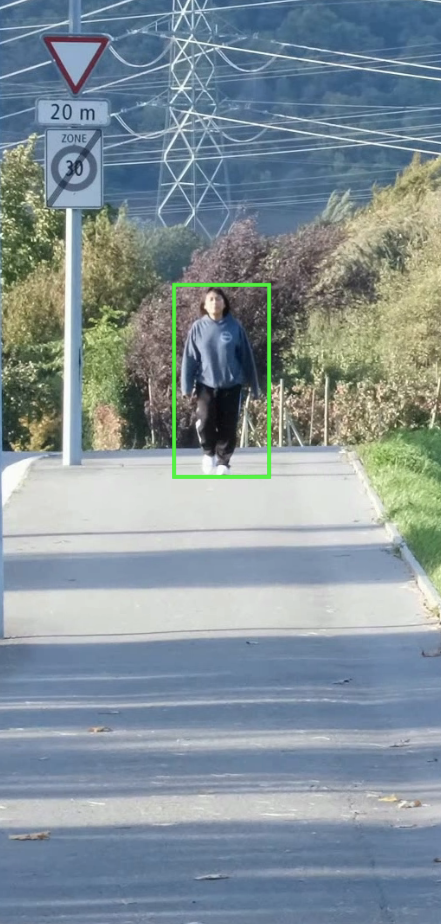}%
        \includegraphics[width=0.125\linewidth]{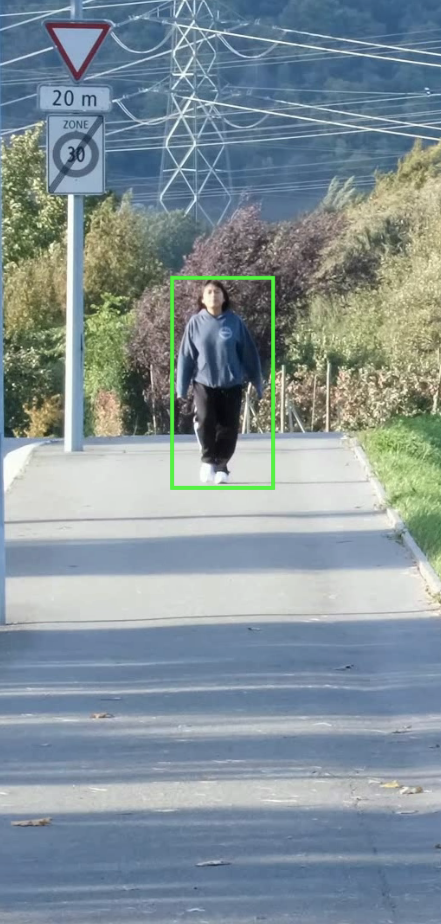}%
        \includegraphics[width=0.125\linewidth]{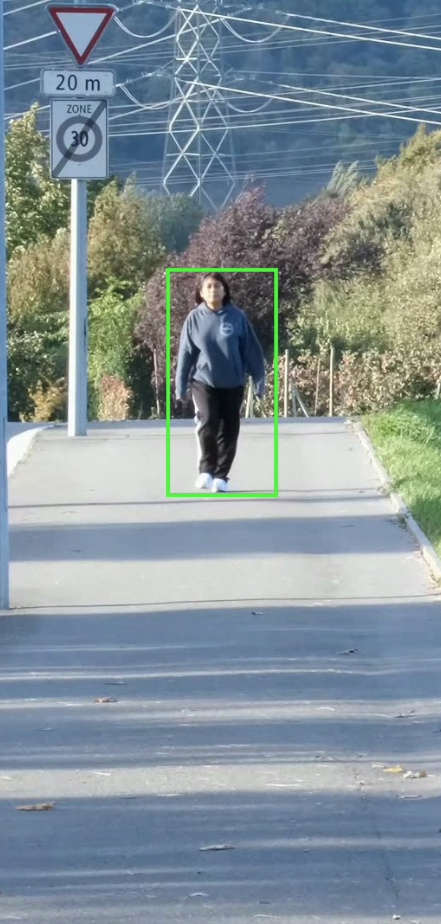}%
        \includegraphics[width=0.125\linewidth]{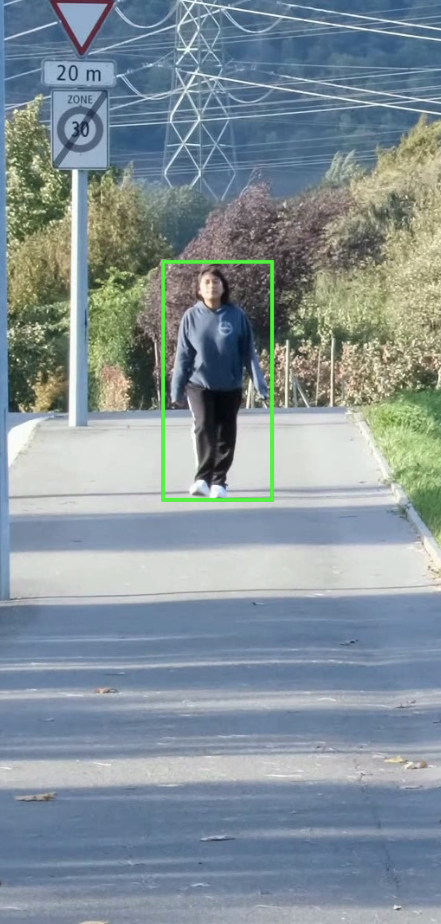}%
        \includegraphics[width=0.125\linewidth]{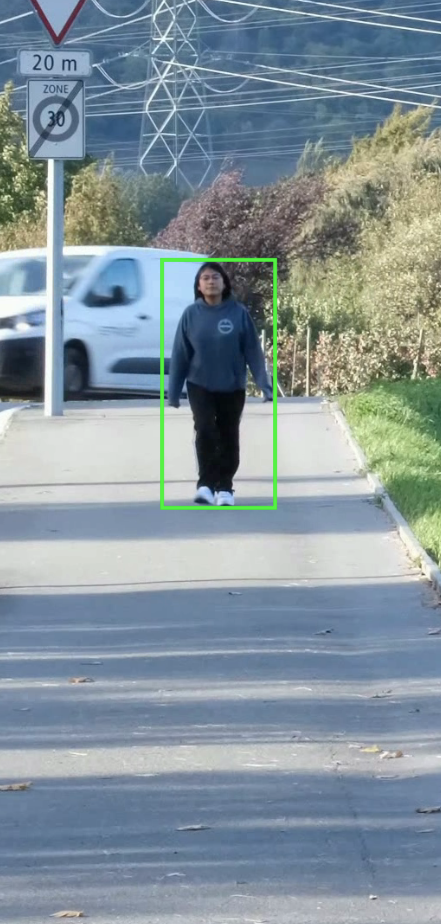}%
        \includegraphics[width=0.125\linewidth]{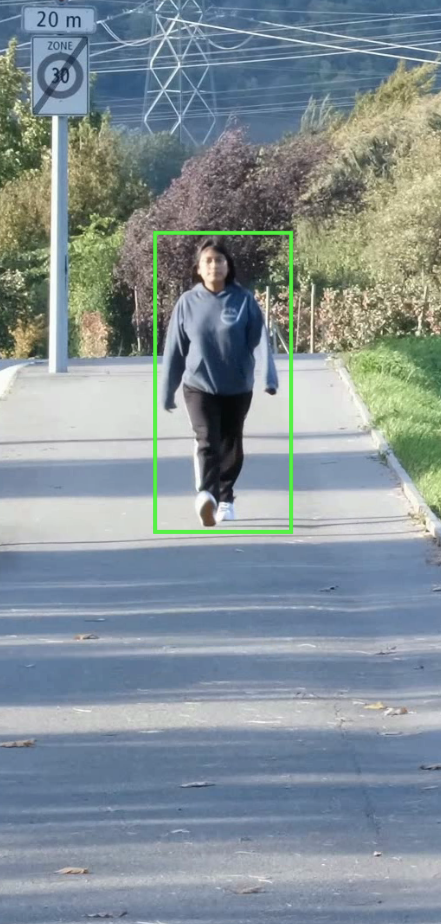}\\%
        \nointerlineskip
        \includegraphics[width=0.125\linewidth]{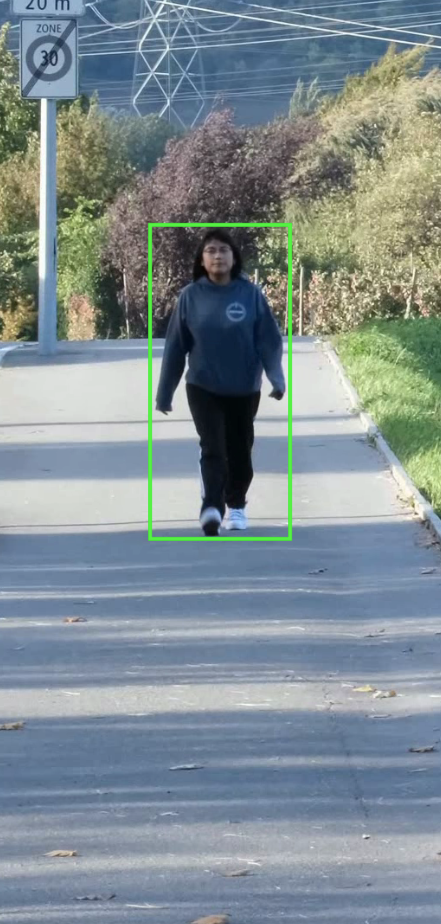}%
        \includegraphics[width=0.125\linewidth]{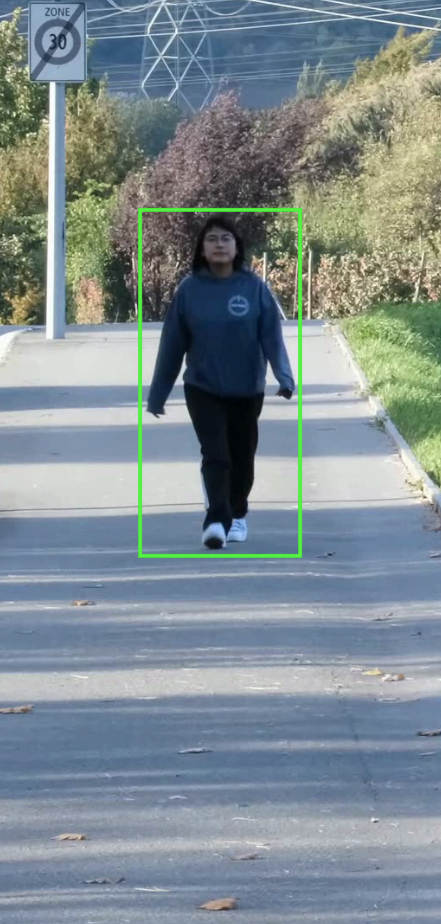}%
        \includegraphics[width=0.125\linewidth]{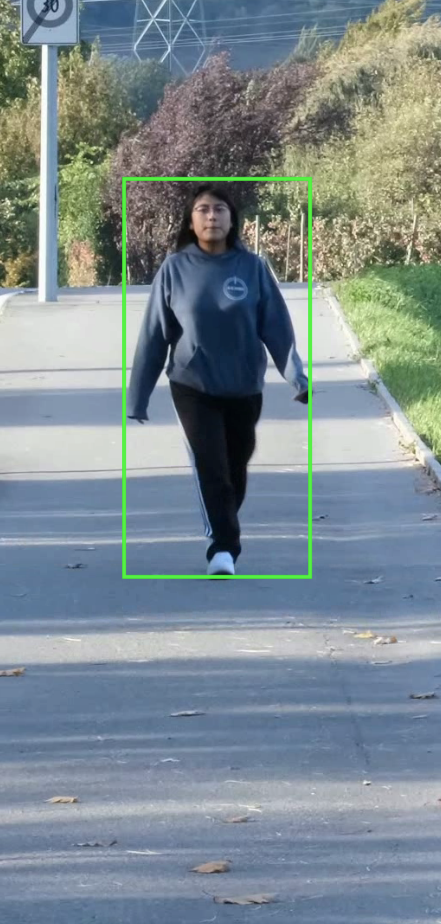}%
        \includegraphics[width=0.125\linewidth]{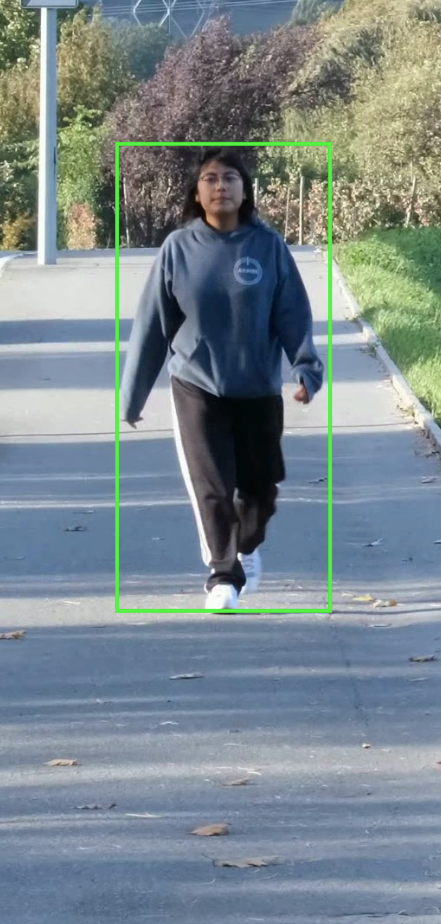}%
        \includegraphics[width=0.125\linewidth]{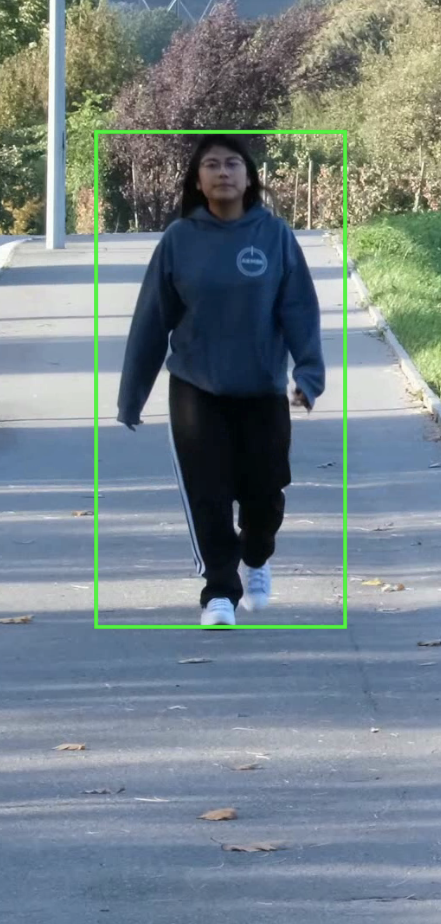}%
        \includegraphics[width=0.125\linewidth]{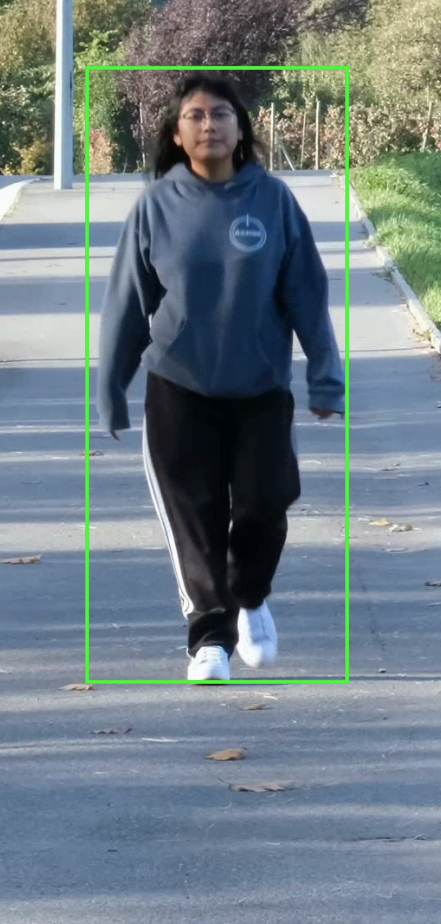}%
        \includegraphics[width=0.125\linewidth]{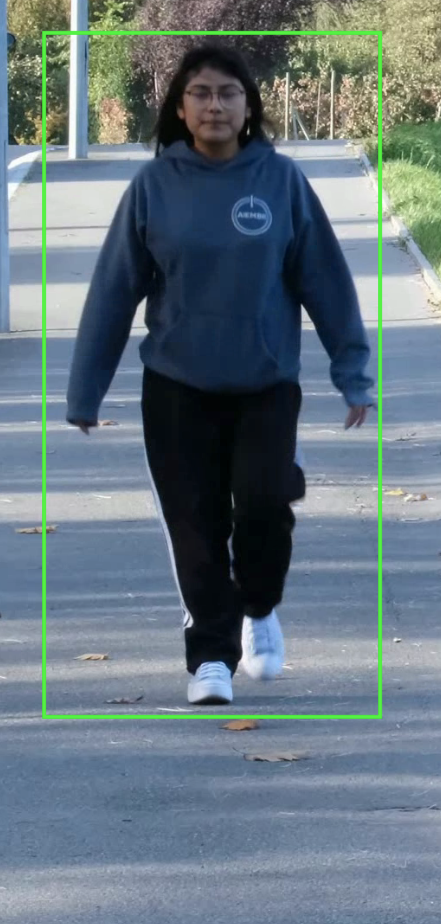}%
        \includegraphics[width=0.125\linewidth]{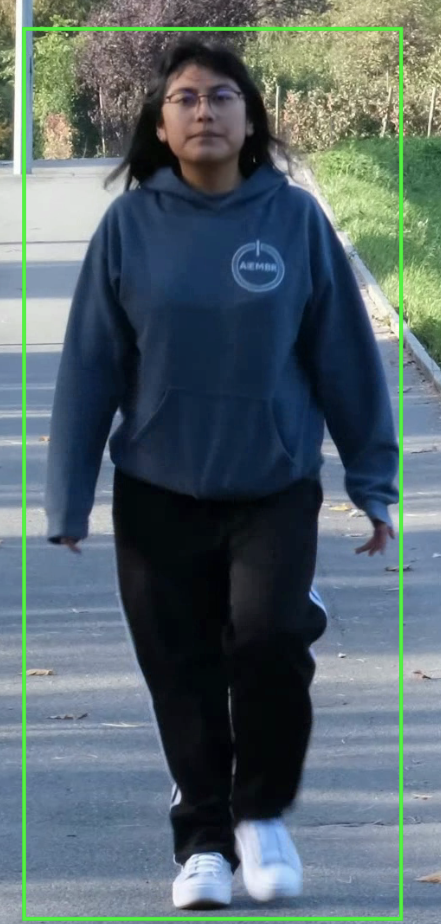}
        \caption{"Walking Towards Camera" (WTC) scenario.}
        \label{subfig:gait_sequence}
    \end{subfigure}

    \caption{Experimental overview: (a) station layout and path; (b) hardware comparison; and (c) sample frames showing long- to near-range capture.}
    \label{fig:front_page_fig}
\end{figure}

%% file: tables/fr_protocols.tex
\begin{table}[hbt!]
\centering
\resizebox{\columnwidth}{!}{%
\begin{tabular}{lll}
\hline
\textbf{Protocol} & \textbf{Enroll (\# Frames, Criteria)} & \textbf{Probe (\# Frames, Criteria)} \\ \hline
\textbf{HQ-[1, 3]} & 1, 3 (Frontal) & 1 (HQ) \\
\textbf{HQ-[10, 20]} & 10, 20 (Mix: Frontal/Tilted) & 1 (HQ) \\
\textbf{GF-3} & 3 (Frontal) & [1, 10, 20, 40] (LQ) \\
\textbf{GF-10} & 10 (Mix: Frontal/Tilted) & [1, 10, 20, 40] (LQ) \\
\textbf{3F} & 3 (Frontal) & [1, 20, 40] (LQ) \\
\textbf{WTC} & 3 (Frontal) & 1 (HQ/LQ) \\ \hline
\end{tabular}%
}
\caption{FR benchmarking protocols. HQ = High Quality (100m zoom), LQ = Low Quality (no zoom), GF = Ground Floor, 3F = 3rd Floor, WTC = Walking Towards Camera. All enrollment data were captured using the frontal cameras of mobile devices, imitating a real pre-enrollment scenario.}
\label{tab:fr_protocols}
\end{table}

%% file: figures/gait_views_variations.tex
\begin{figure}[tbp]
    \centering
    \captionsetup[subfloat]{labelformat=empty}
    
    \subfloat{\includegraphics[width=0.11\columnwidth]{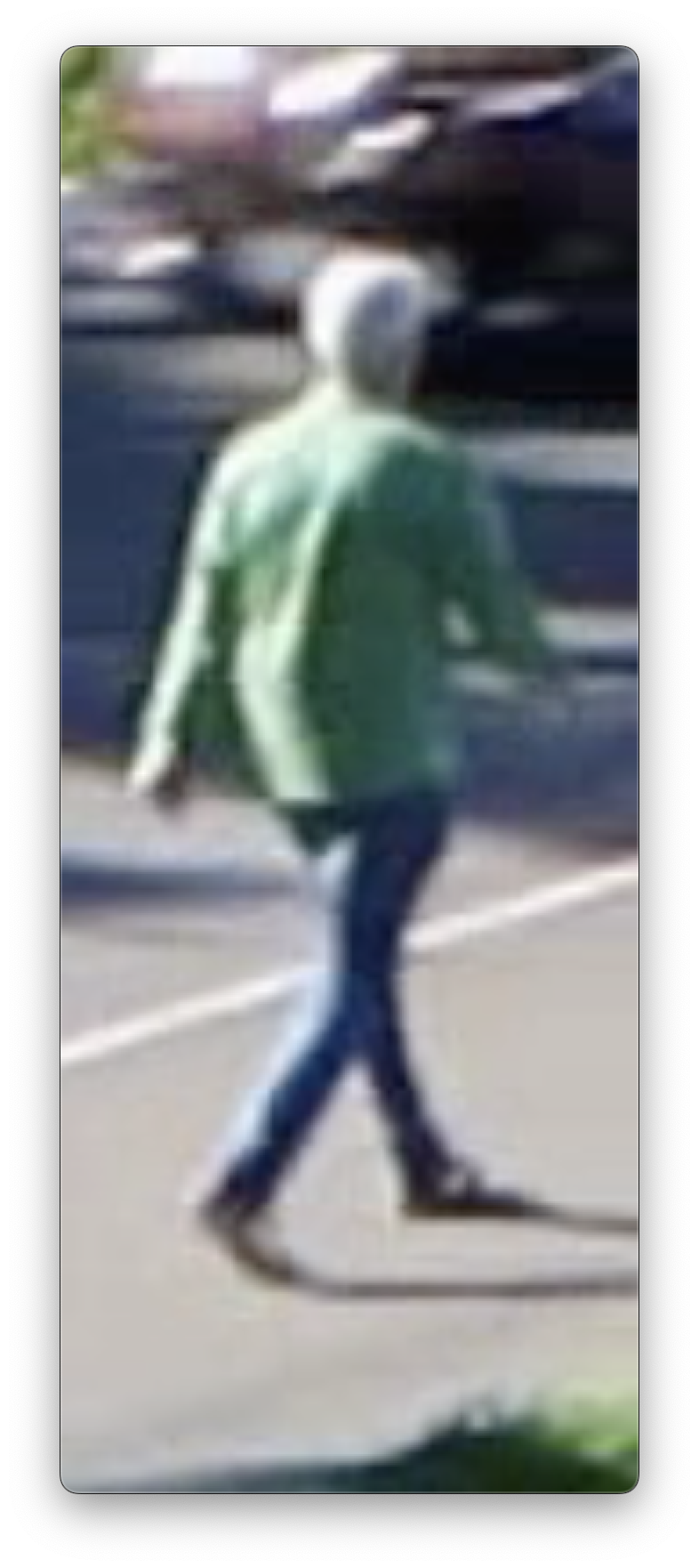}}\hfill
    \subfloat{\includegraphics[width=0.11\columnwidth]{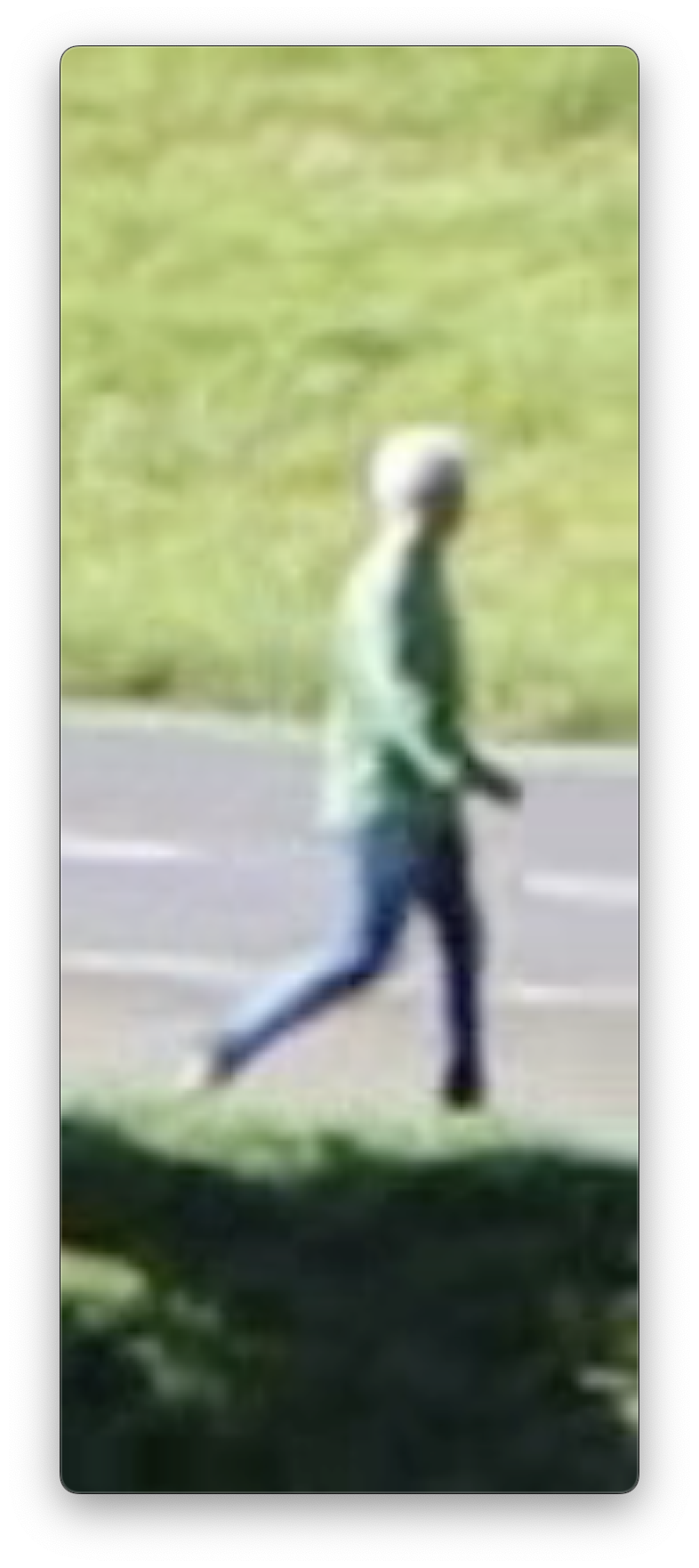}}\hfill
    \subfloat{\includegraphics[width=0.11\columnwidth]{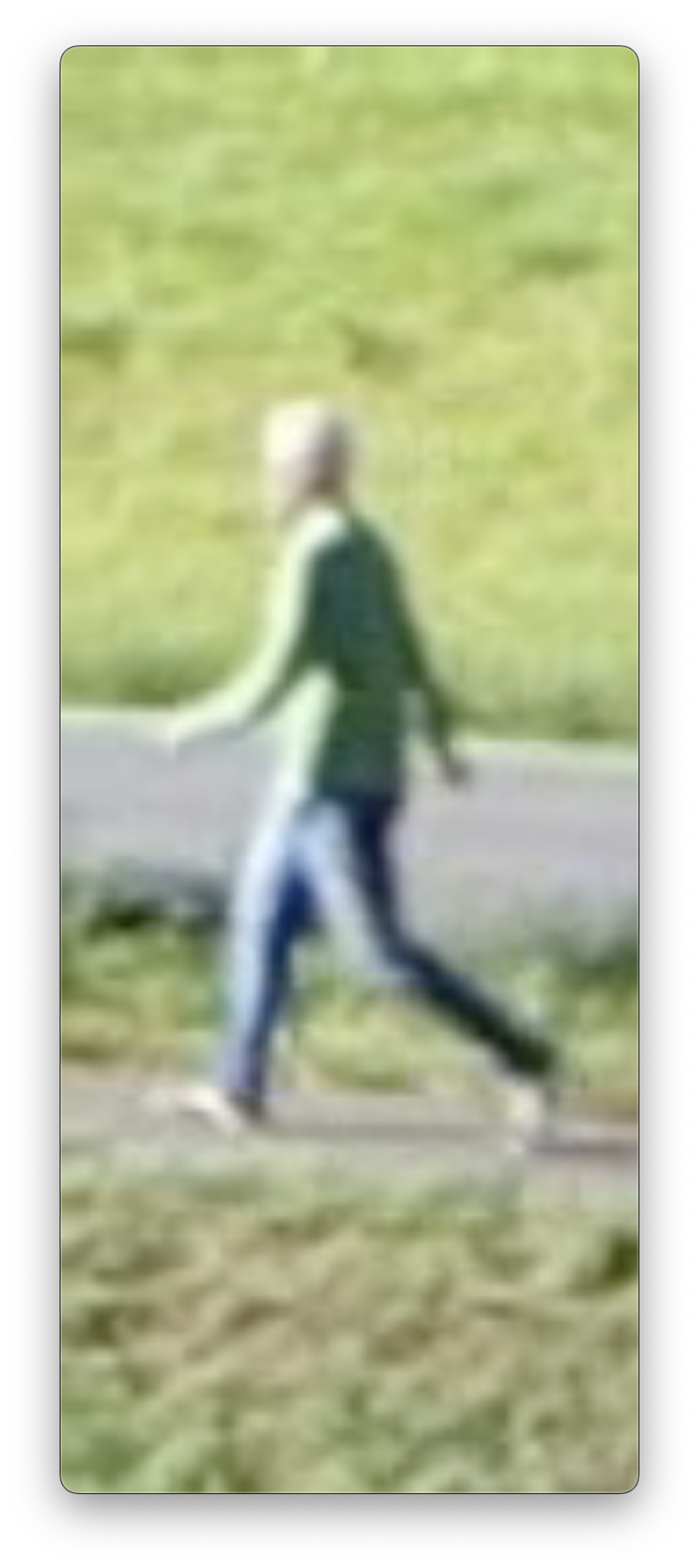}}\hfill
    \subfloat{\includegraphics[width=0.11\columnwidth]{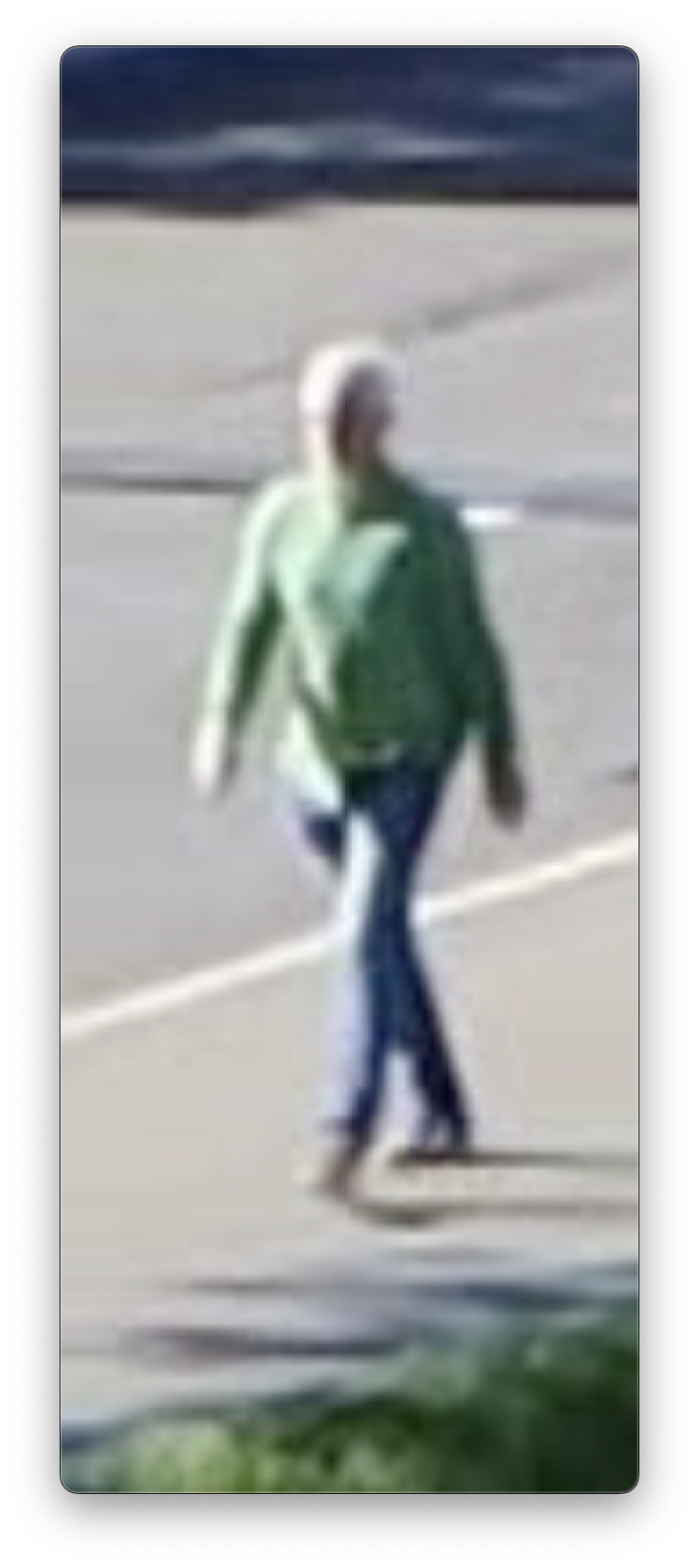}}
    \hspace{0.5em}\vrule\hspace{0.5em}
    \subfloat{\includegraphics[width=0.11\columnwidth]{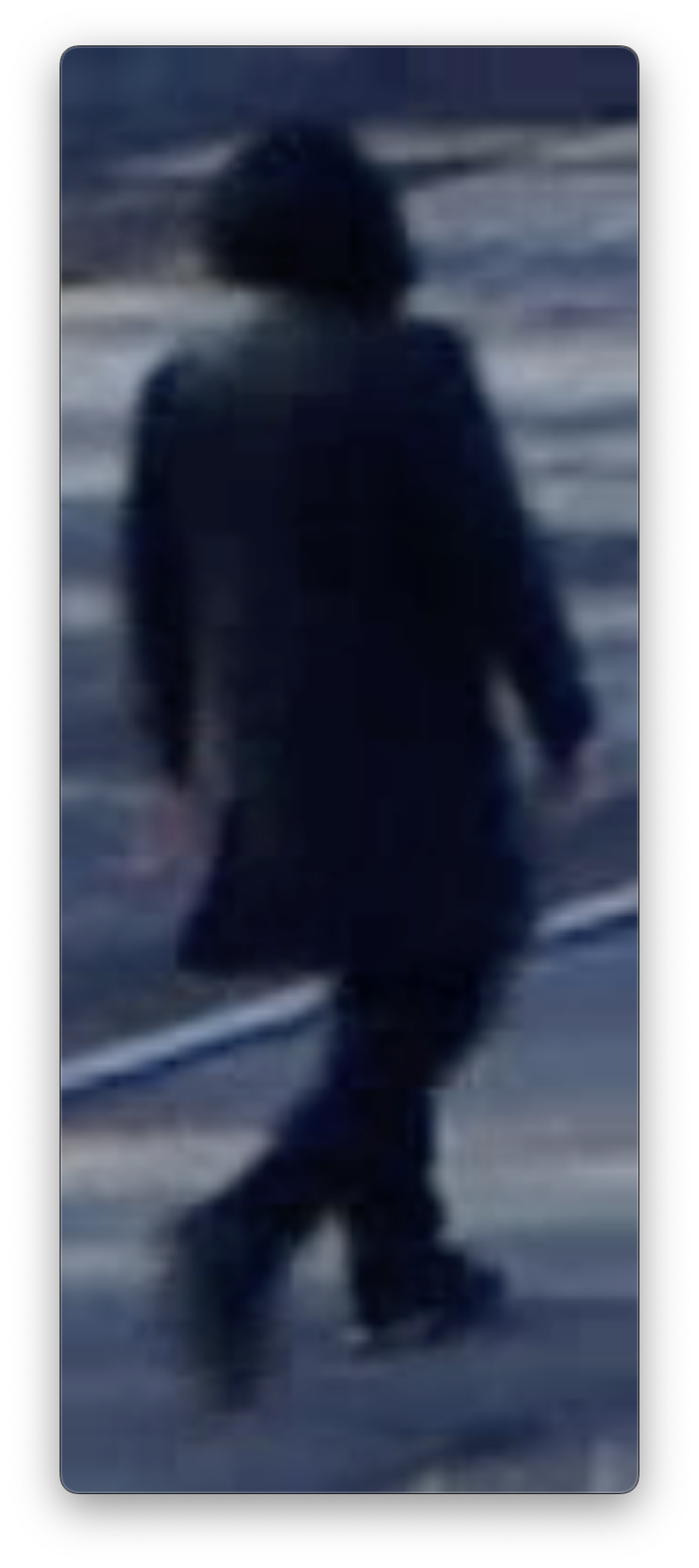}}\hfill
    \subfloat{\includegraphics[width=0.11\columnwidth]{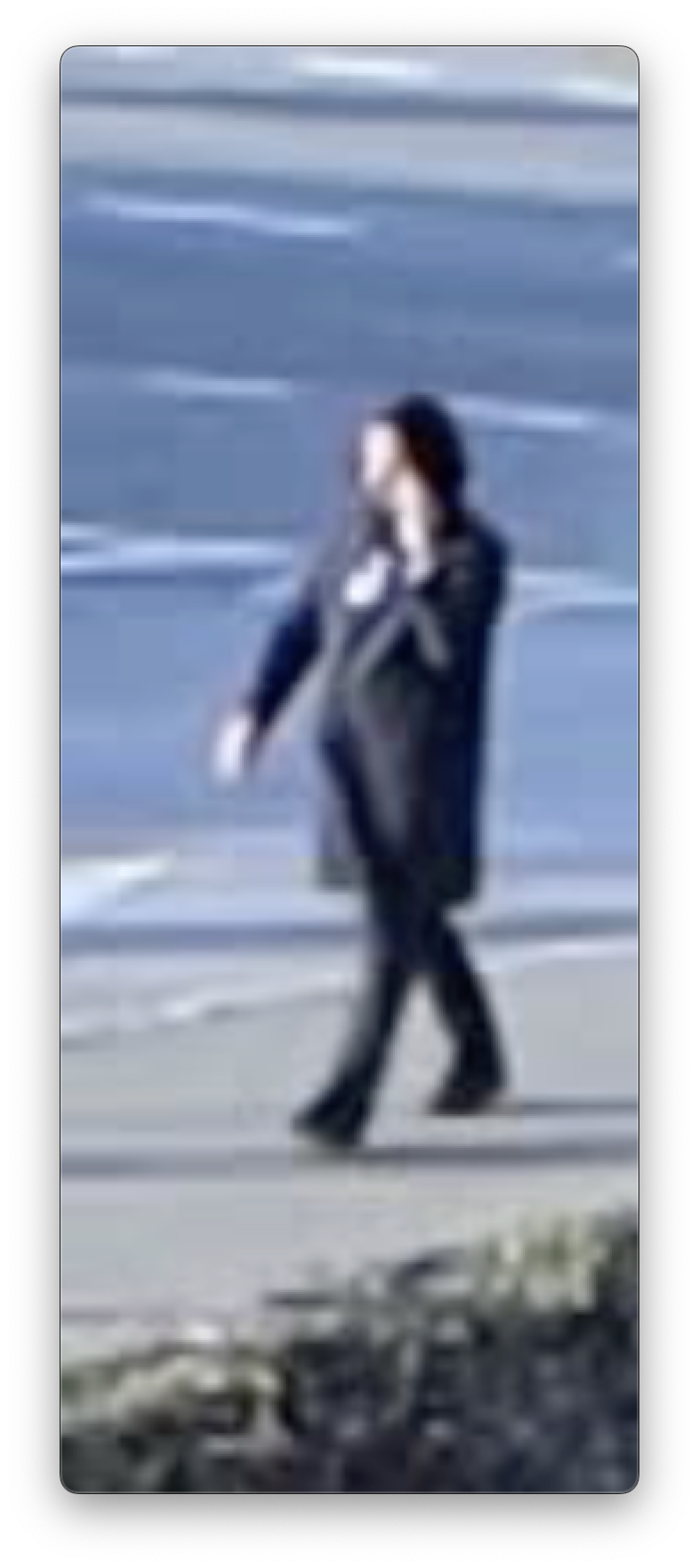}}\hfill
    \subfloat{\includegraphics[width=0.11\columnwidth]{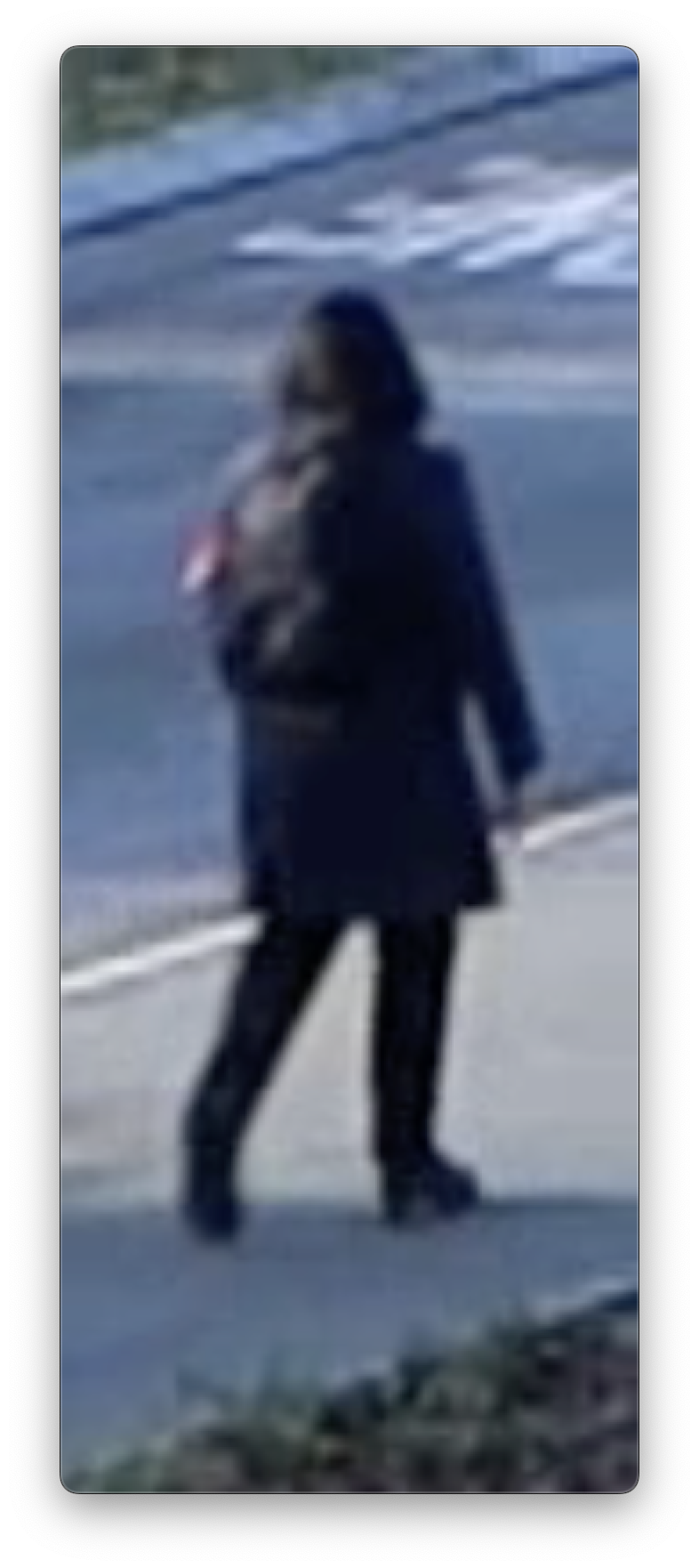}}\hfill
    \subfloat{\includegraphics[width=0.11\columnwidth]{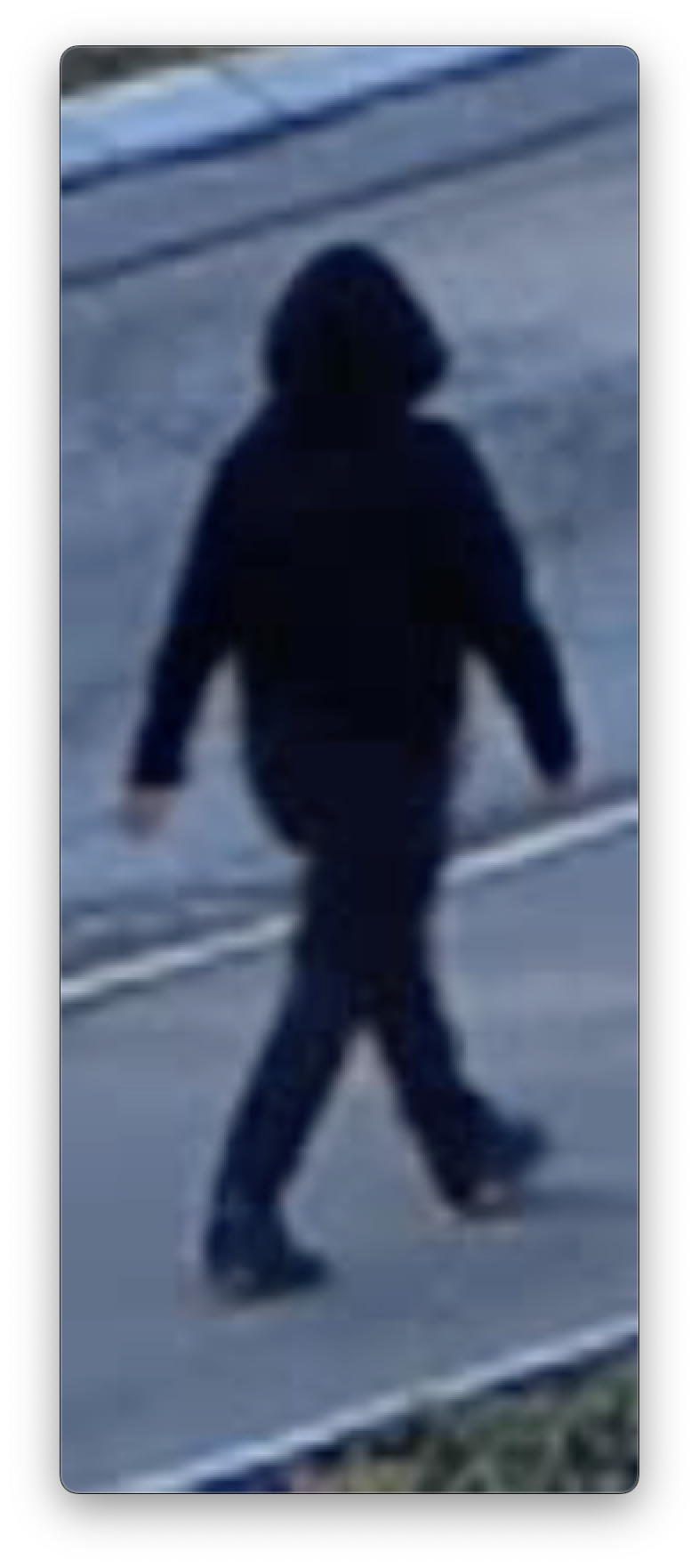}}

    \vspace{0.5em} 

    \subfloat[V1]{\includegraphics[width=0.11\columnwidth]{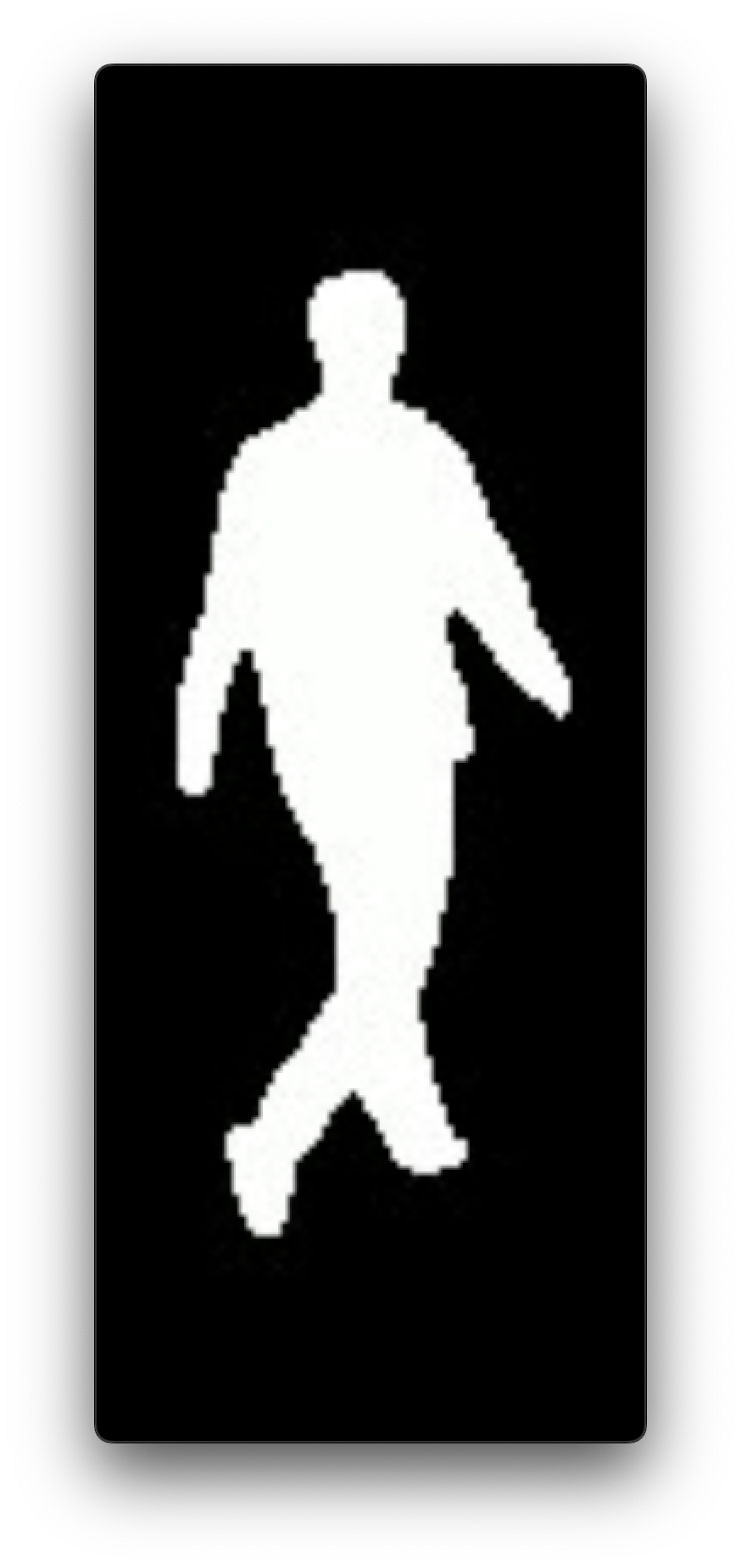}}\hfill
    \subfloat[V2]{\includegraphics[width=0.11\columnwidth]{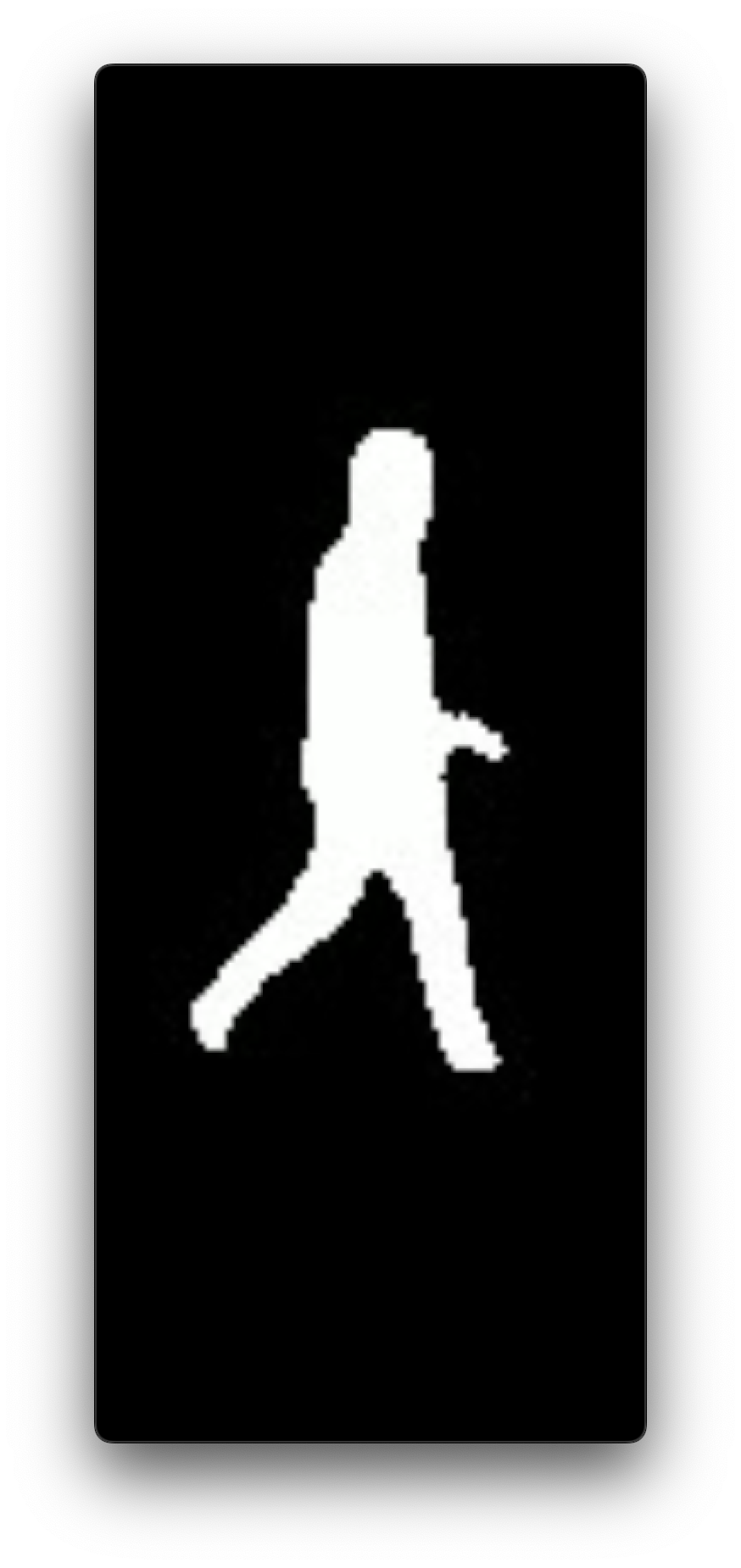}}\hfill
    \subfloat[V3]{\includegraphics[width=0.11\columnwidth]{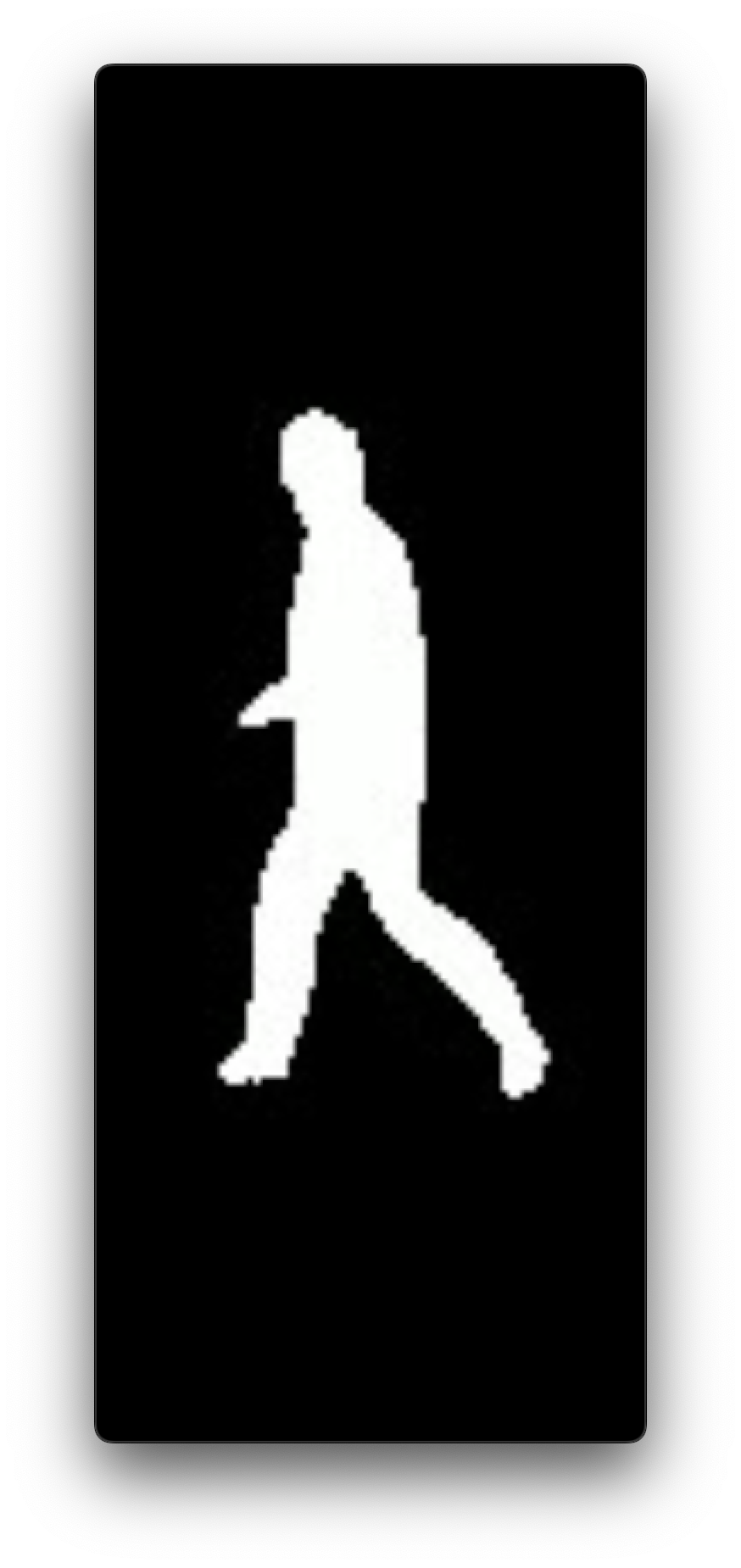}}\hfill
    \subfloat[V4]{\includegraphics[width=0.11\columnwidth]{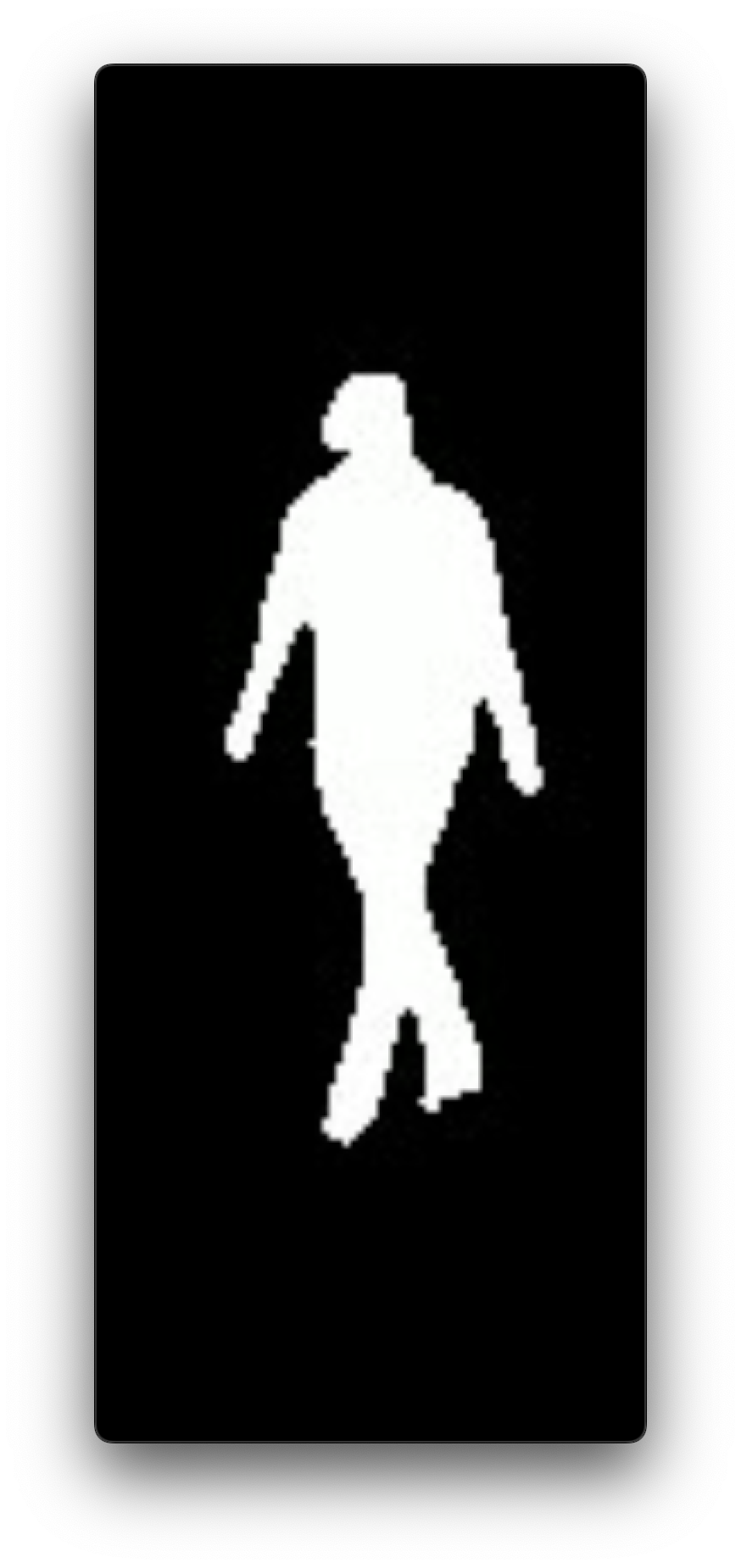}}
    \hspace{0.5em}\vrule\hspace{0.5em}
    \subfloat[Normal]{\includegraphics[width=0.11\columnwidth]{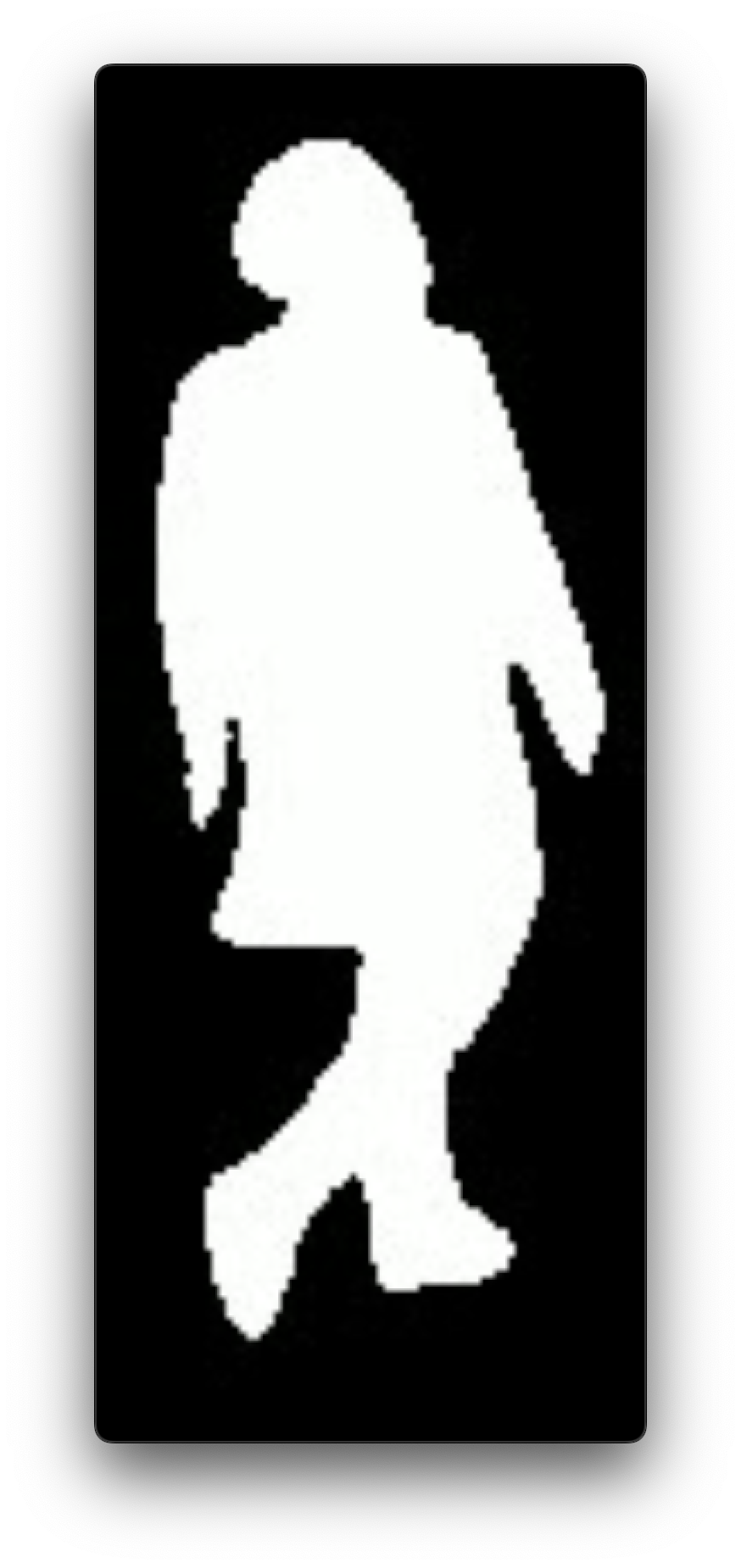}}\hfill
    \subfloat[Phone]{\includegraphics[width=0.11\columnwidth]{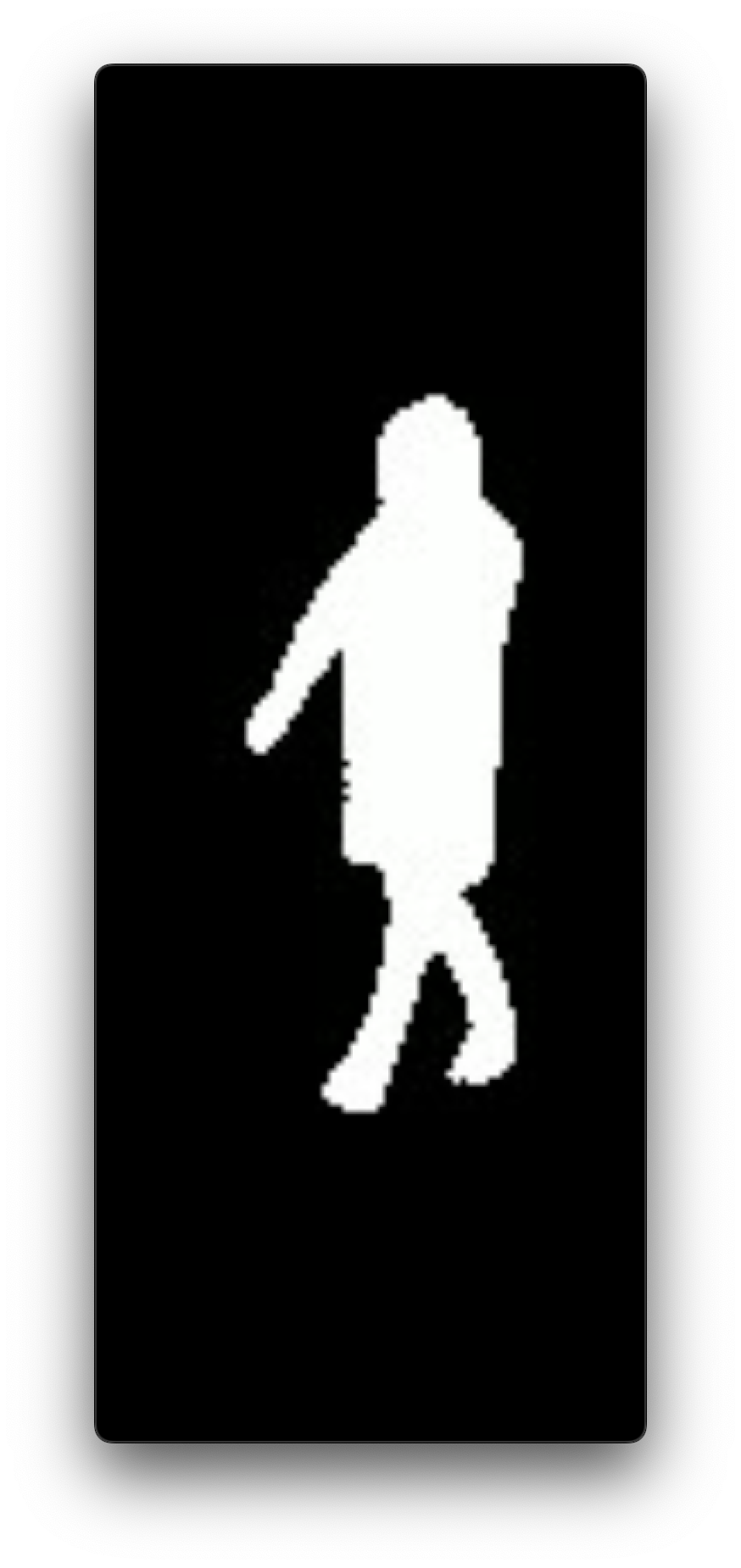}}\hfill
    \subfloat[Bag]{\includegraphics[width=0.11\columnwidth]{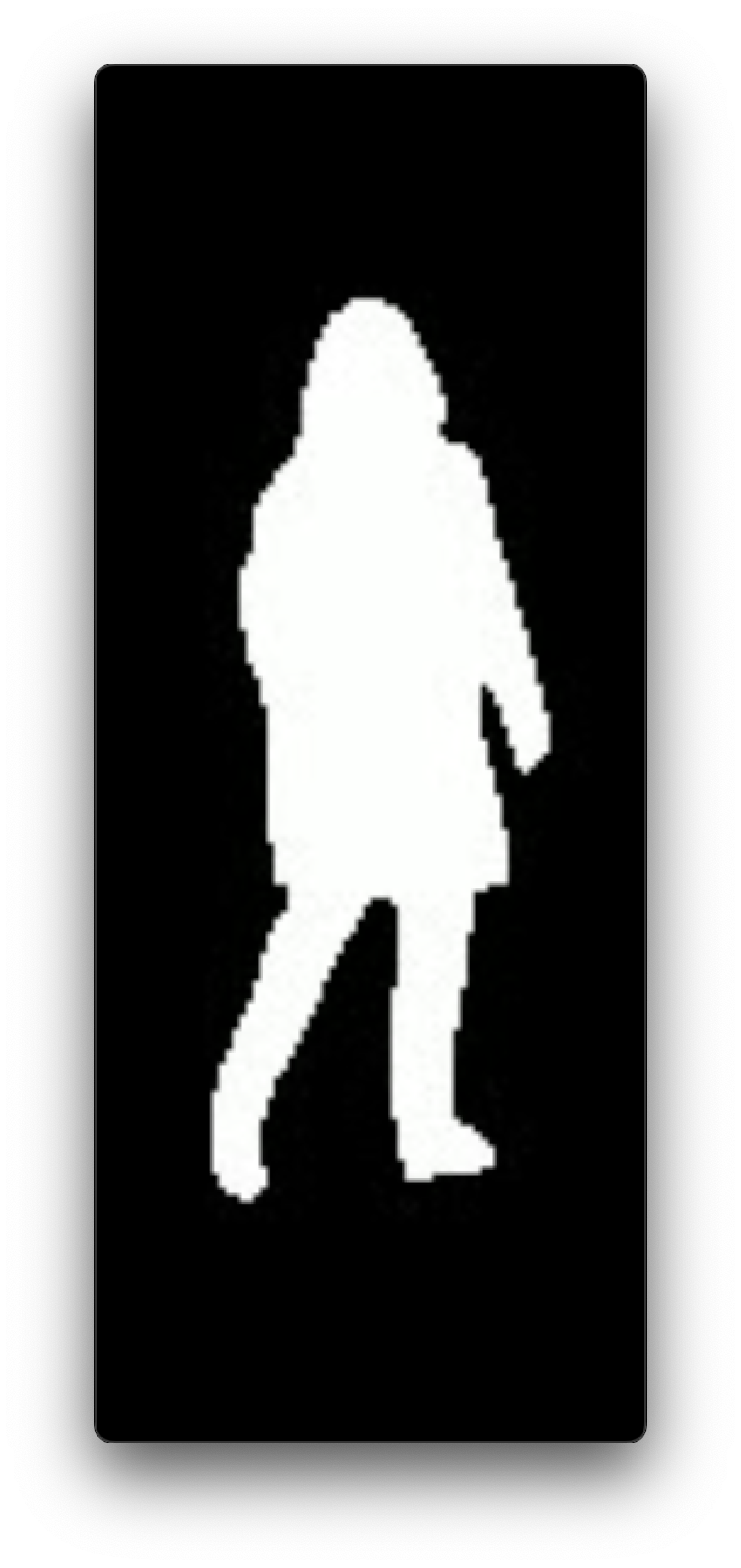}}\hfill
    \subfloat[Clothes]{\includegraphics[width=0.11\columnwidth]{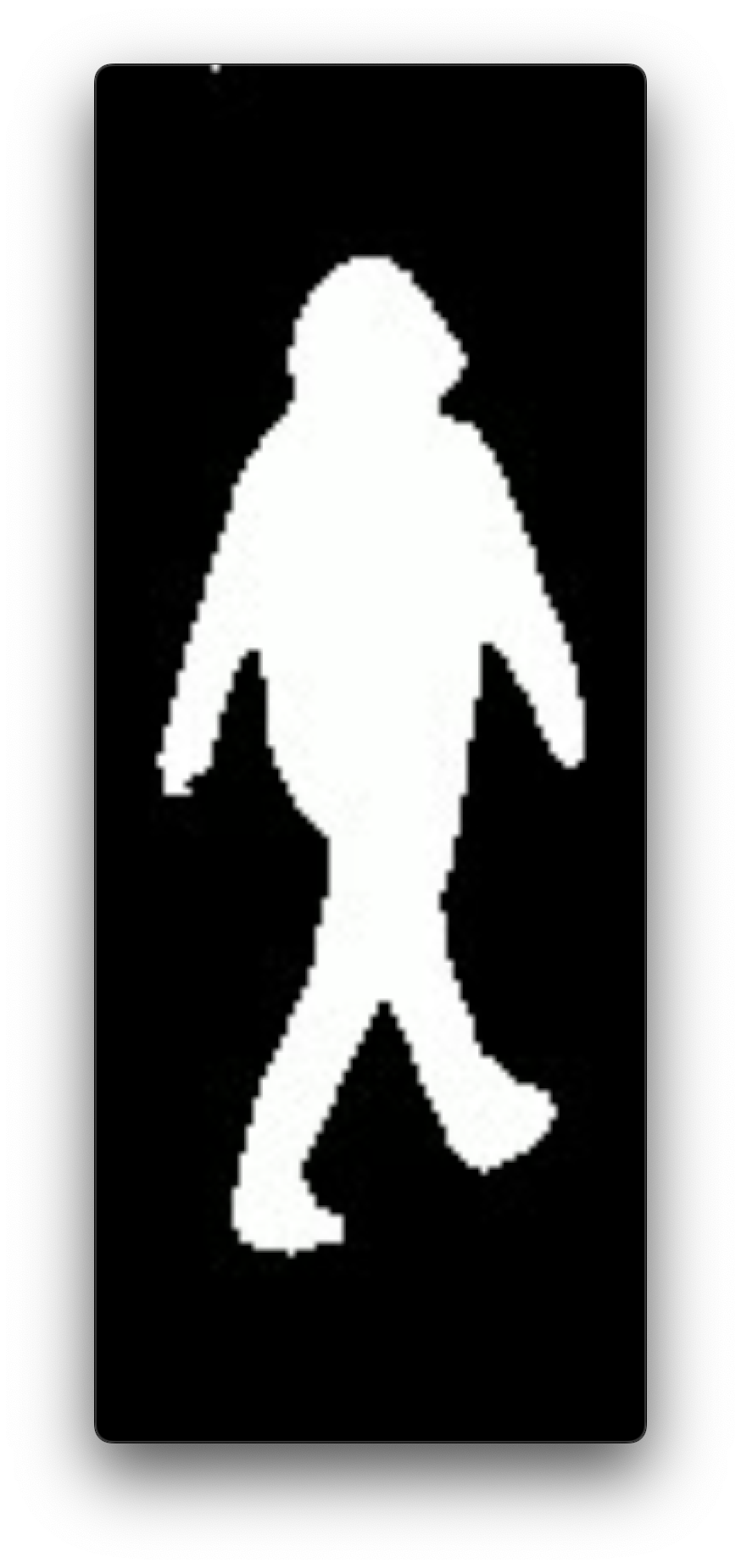}}

    \caption{Gait analysis data: RGB images (top) and generated silhouettes (bottom). Left four columns show walking direction variations; right four columns show walking style and appearance variations.}
    \label{fig:combined_gait_data}
\end{figure}

%% file: tables/fr_sota_models.tex
\begin{table}[t] 
\centering
\resizebox{\columnwidth}{!}{%
\begin{tabular}{llcll}
\toprule
\textbf{Model} & \textbf{Arch.} & \textbf{Params} & \textbf{Innovation} & \textbf{Target} \\ 
\midrule
\textbf{IR50}     & ResNet & 43.6M & Adaptive Margin & Robustness/Latency \\
\textbf{IR101}    & ResNet & 65.2M & Deeper Backbone & High Accuracy \\
\textbf{EdgeFace} & Hybrid & 18.2M  & LoRaLin/STDA Encoder & Edge/IoT \\
\textbf{LV-Face}  & ViT    & 140M+ & Hybrid Loss     & Max Accuracy \\ 
\bottomrule
\end{tabular}%
}
\caption{Comparison of the selected Face Recognition models.}
\label{tab:fr_sota_models}
\end{table}

%% file: tables/sota_fr_results.tex
\begin{table}[t]
\centering
\resizebox{\columnwidth}{!}{%
\begin{tabular}{lcccccccc}
\toprule
\multirow{2}{*}{\textbf{Model}} & \multicolumn{2}{c}{\textbf{HQ-1}} & \multicolumn{2}{c}{\textbf{HQ-3}} & \multicolumn{2}{c}{\textbf{HQ-10}} & \multicolumn{2}{c}{\textbf{HQ-20}} \\
\cmidrule(lr){2-3} \cmidrule(lr){4-5} \cmidrule(lr){6-7} \cmidrule(lr){8-9}
 & \textbf{1\%} & \textbf{0.1\%} & \textbf{1\%} & \textbf{0.1\%} & \textbf{1\%} & \textbf{0.1\%} & \textbf{1\%} & \textbf{0.1\%} \\
\midrule
IR50     & 1.3 & 1.8 & 1.1 & 1.6 & 0.9 & 1.2 & 0.9 & 1.1 \\
IR101    & \textbf{0.9} & 1.3 & 0.9 & 1.1 & 0.8 & 1.0 & 0.8 & 0.9 \\
EdgeFace & 1.8 & 2.5 & 1.5 & 2.1 & 1.3 & 1.8 & 1.3 & 1.8 \\
LVFace   & \textbf{0.9} & \textbf{1.2} & \textbf{0.5} & \textbf{0.7} & \textbf{0.4} & \textbf{0.6} & \textbf{0.4} & \textbf{0.6} \\
\bottomrule
\end{tabular}%
}
\caption{FNMR (\%) evaluation results for HQ protocols. Model thresholds were calibrated on IJB-C at FMR = 1\% and 0.1\%.}
\label{tab:fr_HQ_results}
\end{table}

%% file: figures/cmc_plots/cmc_g1.tex
\begin{figure}[t]
    \centering
    \includegraphics[width=\linewidth]{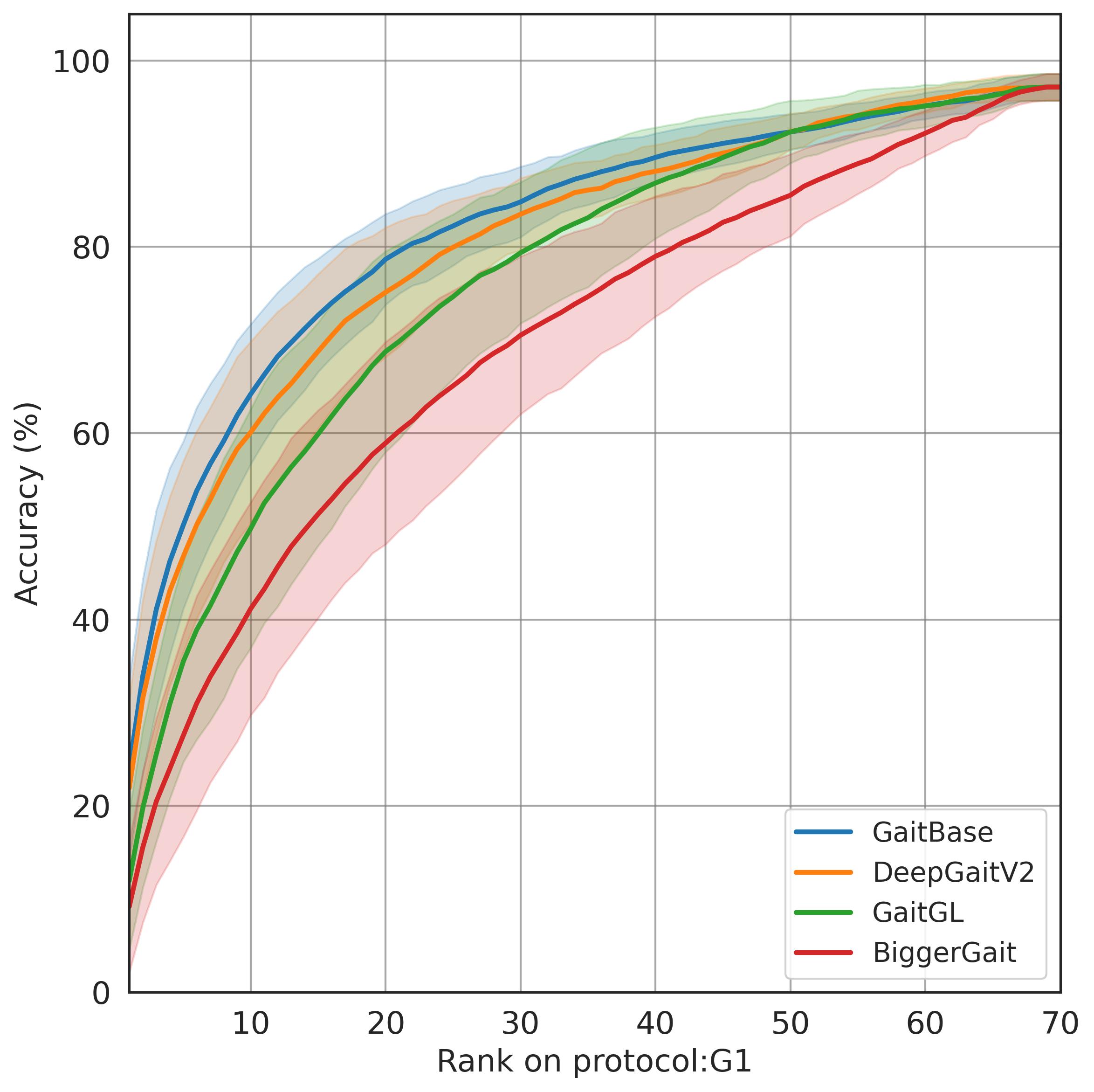}
    \caption{CMC plot for gait recognition protocol G1: Here GaitBase~\cite{fan2023opengait} performs the best.}
    \label{fig:cmc_g1}
\end{figure}

%% file: tables/silhouette_res.tex
\begin{table*}[t]
\centering
\resizebox{\textwidth}{!}{%
\begin{tabular}{lllccccc|ccccc}
\toprule
& & & \multicolumn{5}{c|}{\textbf{Rank-1 Acc. on G1}} & \multicolumn{5}{c}{\textbf{Rank-1 Acc. on G2}} \\
\cmidrule(lr){4-8} \cmidrule(lr){9-13}
\textbf{Method} & \textbf{Train} & \textbf{Condition} & \textbf{V1} & \textbf{V2} & \textbf{V3} & \textbf{V4} & \textbf{Mean} & \textbf{V1} & \textbf{V2} & \textbf{V3} & \textbf{V4} & \textbf{Mean} \\
\midrule
\multirow{4}{*}{GaitBase}   & \multirow{4}{*}{GREW}
  & Normal  & 28.10 & 20.10 & 29.05 & 29.95 & \textbf{26.80} & 14.36 & 26.15 & 25.64 & 17.95 & 21.02 \\
& & Phone   & 20.00 & 17.65 & 23.33 & 26.57 & 21.89 &  8.99 & 27.96 & 24.87 & 13.23 & 18.76 \\
& & Bag     & 25.24 & 18.57 & 20.95 & 28.57 & \textbf{23.33} & 11.11 & 25.25 & 25.25 & 18.18 & 19.95 \\
& & Clothes & 23.67 & 24.15 & 21.26 & 27.05 & \textbf{24.03} &  7.58 & 20.00 & 23.74 & 12.12 & 15.86 \\
\midrule
\multirow{4}{*}{DeepGaitV2} & \multirow{4}{*}{GREW}
  & Normal  & 27.62 & 19.12 & 21.91 & 29.47 & 24.53 & 25.13 & 28.21 & 36.92 & 24.62 & \textbf{28.72} \\
& & Phone   & 27.14 & 15.69 & 19.52 & 28.02 & \textbf{22.59} & 19.05 & 27.96 & 31.75 & 27.52 & \textbf{26.57} \\
& & Bag     & 22.86 & 14.29 & 14.76 & 26.19 & 19.52 & 21.72 & 24.24 & 34.34 & 19.70 & \textbf{25.00} \\
& & Clothes & 24.64 & 15.46 & 15.46 & 28.50 & 21.01 & 19.19 & 26.15 & 29.29 & 20.71 & \textbf{23.84} \\
\midrule
\multirow{4}{*}{GaitGL}     & \multirow{4}{*}{CASIA-B}
  & Normal  & 10.47 &  8.33 & 15.72 & 14.49 & 12.25 &  5.13 & 14.36 & 11.80 &  2.57 &  8.46 \\
& & Phone   & 10.00 &  9.80 & 17.14 & 13.04 & 12.50 &  4.76 &  9.68 &  8.99 &  4.76 &  7.05 \\
& & Bag     &  9.52 &  7.14 & 13.81 & 13.81 & 11.07 &  3.03 & 11.62 & 10.61 &  3.03 &  7.07 \\
& & Clothes &  9.66 &  9.18 & 14.49 & 14.98 & 12.08 &  5.05 & 11.28 &  8.59 &  4.55 &  7.37 \\
\midrule
\multirow{4}{*}{BiggerGait} & \multirow{4}{*}{CCPG}
  & Normal  &  6.67 &  6.86 & 11.43 & 10.14 &  8.78 &  1.93 &  3.98 &  3.92 &  2.90 &  3.18 \\
& & Phone   & 10.48 &  2.94 &  8.57 &  8.70 &  7.67 &  1.99 &  2.53 &  3.98 &  3.49 &  3.00 \\
& & Bag     &  5.24 &  8.09 & 13.33 & 11.90 &  9.64 &  2.94 &  3.92 &  5.39 &  1.96 &  3.55 \\
& & Clothes & 10.63 &  9.18 & 11.90 & 10.95 & 10.67 &  1.93 &  5.32 &  4.83 &  3.38 &  3.86 \\
\bottomrule
\end{tabular}}
\caption{Rank-1 accuracy (\%) on G1 and G2 gait protocol.
Probe sequences: Normal, Phone, Bag, Clothes. Gallery: Normal.
V1-V4 denote the four camera views. GaitBase~\cite{fan2023opengait} (88M) and DeepGaitV2~\cite{fan2023exploring} (95M) are trained on GREW~\cite{guo2025gait},
GaitGL~\cite{lin2022gaitgl}(3.1M) on CASIA-B~\cite{casiab}, and BiggerGait~\cite{ye2025biggergait}(140M) on CCPG~\cite{ccpg}.}
\label{tab:gait_all_results}
\end{table*}

%% file: figures/cmc_plots/cmc_g2.tex
\begin{figure}[t]
    \centering
    \includegraphics[width=\linewidth]{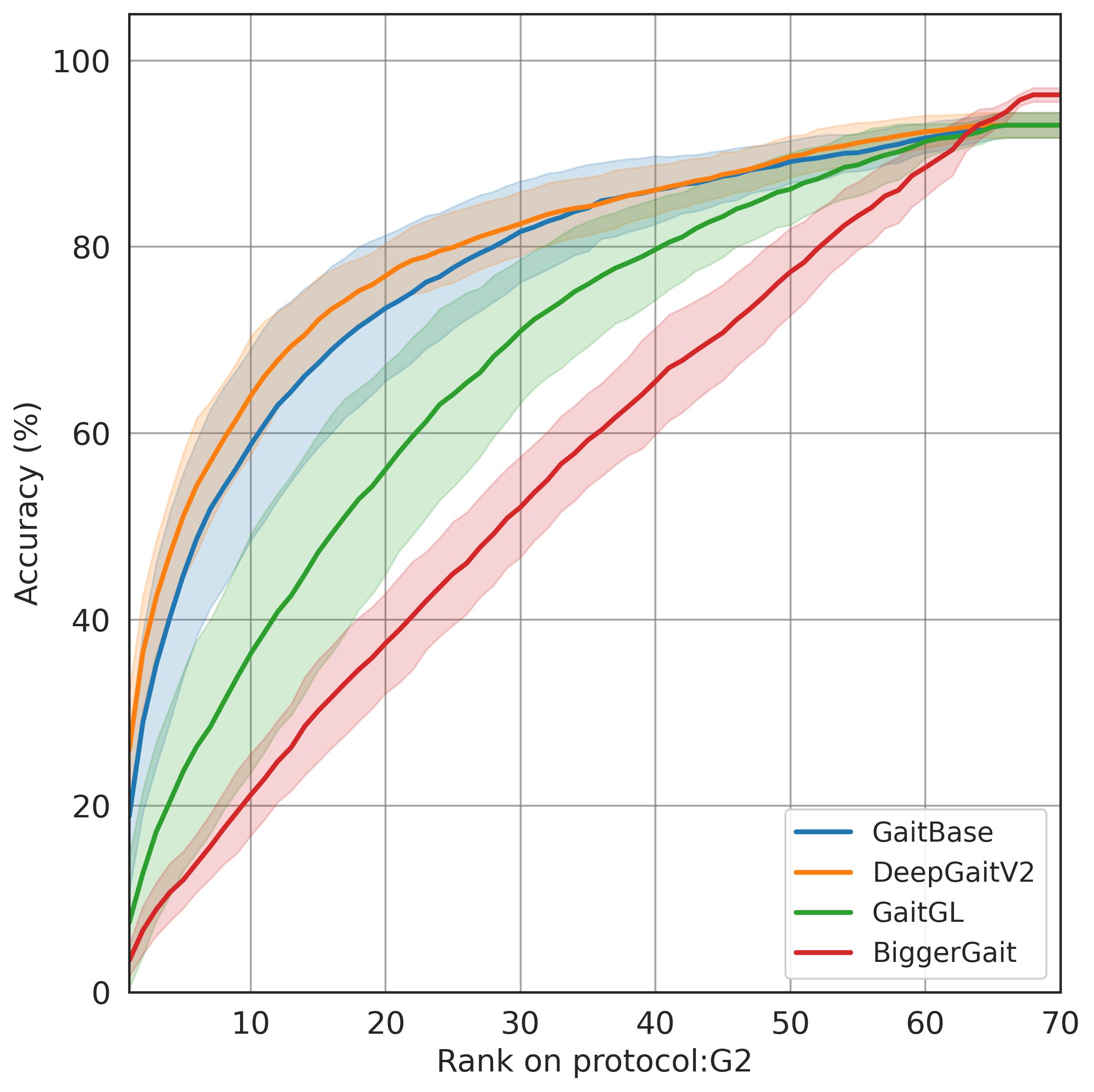}
    \caption{CMC plot for gait recognition protocol G2: Here DeepGaitV2~\cite{fan2023exploring} performs the best. This protocol is difficult than G1 as there is an extreme viewpoint change between two session.}
    \label{fig:cmc_g2}
\end{figure}

%% file: egbib.bib
@inproceedings{fan2023opengait,
  title={Opengait: Revisiting gait recognition towards better practicality},
  author={Fan, Chao and Liang, Junhao and Shen, Chuanfu and Hou, Saihui and Huang, Yongzhen and Yu, Shiqi},
  booktitle={Proceedings of the IEEE/CVF conference on computer vision and pattern recognition},
  pages={9707--9716},
  year={2023}
}

@article{lin2022gaitgl,
  title={Gaitgl: Learning discriminative global-local feature representations for gait recognition},
  author={Lin, Beibei and Zhang, Shunli and Wang, Ming and Li, Lincheng and Yu, Xin},
  journal={arXiv preprint arXiv:2208.01380},
  year={2022}
}

@article{ye2025biggergait,
  title={Biggergait: Unlocking gait recognition with layer-wise representations from large vision models},
  author={Ye, Dingqiang and Fan, Chao and Huang, Zhanbo and Luo, Chengwen and Li, Jianqiang and Yu, Shiqi and Liu, Xiaoming},
  journal={arXiv preprint arXiv:2505.18132},
  year={2025}
}

@InProceedings{ccpg,
    author    = {Li, Weijia and Hou, Saihui and Zhang, Chunjie and Cao, Chunshui and Liu, Xu and Huang, Yongzhen and Zhao, Yao},
    title     = {An In-Depth Exploration of Person Re-Identification and Gait Recognition in Cloth-Changing Conditions},
    booktitle = {Proceedings of the IEEE/CVF Conference on Computer Vision and Pattern Recognition (CVPR)},
    month     = {June},
    year      = {2023},
    pages     = {13824-13833}
}

@inproceedings{casiab,
  title={A framework for evaluating the effect of view angle, clothing and carrying condition on gait recognition},
  author={Yu, Shiqi and Tan, Daoliang and Tan, Tieniu},
  booktitle={18th international conference on pattern recognition (ICPR'06)},
  volume={4},
  pages={441--444},
  year={2006},
  organization={IEEE}
}

@article{oquab2023dinov2,
  title={Dinov2: Learning robust visual features without supervision},
  author={Oquab, Maxime and Darcet, Timoth{\'e}e and Moutakanni, Th{\'e}o and Vo, Huy and Szafraniec, Marc and Khalidov, Vasil and Fernandez, Pierre and Haziza, Daniel and Massa, Francisco and El-Nouby, Alaaeldin and others},
  journal={arXiv preprint arXiv:2304.07193},
  year={2023}
}

@inproceedings{lfw,
  TITLE = {{Labeled Faces in the Wild: A Database forStudying Face Recognition in Unconstrained Environments}},
  AUTHOR = {Huang, Gary B. and Mattar, Marwan and Berg, Tamara and Learned-Miller, Eric},
  URL = {https://inria.hal.science/inria-00321923},
  BOOKTITLE = {{Workshop on Faces in 'Real-Life' Images: Detection, Alignment, and Recognition}},
  ADDRESS = {Marseille, France},
  ORGANIZATION = {{Erik Learned-Miller and Andras Ferencz and Fr{\'e}d{\'e}ric Jurie}},
  YEAR = {2008},
  MONTH = Oct,
  HAL_ID = {inria-00321923},
  HAL_VERSION = {v1},
}

@article{casia-webface,
  title={Learning Face Representation from Scratch},
  author={Dong Yi and Zhen Lei and Shengcai Liao and S. Li},
  journal={ArXiv},
  year={2014},
  volume={abs/1411.7923},
  url={https://api.semanticscholar.org/CorpusID:17188384}}

@article{msceleb1m,
  title={MS-Celeb-1M: A Dataset and Benchmark for Large-Scale Face Recognition},
  author={Yandong Guo and Lei Zhang and Yuxiao Hu and Xiaodong He and Jianfeng Gao},
  journal={ArXiv},
  year={2016},
  volume={abs/1607.08221},
  url={https://api.semanticscholar.org/CorpusID:2908606}
}

@INPROCEEDINGS{partialfc-glint360k,
  author={An, Xiang and Zhu, Xuhan and Gao, Yuan and Xiao, Yang and Zhao, Yongle and Feng, Ziyong and Wu, Lan and Qin, Bin and Zhang, Ming and Zhang, Debing and Fu, Ying},
  booktitle={2021 IEEE/CVF International Conference on Computer Vision Workshops (ICCVW)}, 
  title={Partial FC: Training 10 Million Identities on a Single Machine}, 
  year={2021},
  volume={},
  number={},
  pages={1445-1449},}

@article{ijbs,
  title={IJB-S: IARPA Janus Surveillance Video Benchmark},
  author={Nathan D. Kalka and Brianna Maze and James A. Duncan and Kevin OrConnor and Stephen Elliott and Kaleb Hebert and Julia Bryan and Anil K. Jain},
  journal={2018 IEEE 9th International Conference on Biometrics Theory, Applications and Systems (BTAS)},
  year={2018},
  pages={1-9}
}

@INPROCEEDINGS{briar,
  author={Cornett, David and Brogan, Joel and Barber, Nell and Aykac, Deniz and Baird, Seth and Burchfield, Nick and Dukes, Carl and Duncan, Andrew and Ferrell, Regina and Goddard, Jim and Jager, Gavin and Larson, Matt and Murphy, Bart and Johnson, Christi and Shelley, Ian and Srinivas, Nisha and Stockwell, Brandon and Thompson, Leanne and Yohe, Matt and Zhang, Robert and Dolvin, Scott and Santos-Villalobos, Hector J. and Bolme, David S.},
  booktitle={2023 IEEE/CVF Winter Conference on Applications of Computer Vision Workshops (WACVW)}, 
  title={Expanding Accurate Person Recognition to New Altitudes and Ranges: The BRIAR Dataset}, 
  year={2023},
  volume={},
  number={},
  pages={593-602},
  doi={10.1109/WACVW58289.2023.00066}}

@inproceedings{zhu2021webface260m,
  title={Webface260m: A benchmark unveiling the power of million-scale deep face recognition},
  author={Zhu, Zheng and Huang, Guan and Deng, Jiankang and Ye, Yun and Huang, Junjie and Chen, Xinze and Zhu, Jiagang and Yang, Tian and Lu, Jiwen and Du, Dalong and others},
  booktitle={Proceedings of the IEEE/CVF Conference on Computer Vision and Pattern Recognition},
  pages={10492--10502},
  year={2021}
}

@inproceedings{ijbc,
  title={IARPA Janus Benchmark-C: Face Dataset and Protocol},
  author={Maze, B. and others},
  booktitle={2018 International Conference on Biometrics (ICB)},
  year={2018}
}

@article{scface,
author = {Grgic, Mislav and Delac, Kresimir and Grgic, Sonja},
title = {SCface --- surveillance cameras face database},
year = {2011},
issue_date = {February  2011},
publisher = {Kluwer Academic Publishers},
address = {USA},
volume = {51},
number = {3},
issn = {1380-7501},
journal = {Multimedia Tools Appl.},
url = {https://doi.org/10.1007/s11042-009-0417-2},
doi = {10.1007/s11042-009-0417-2},
month = feb,
pages = {863–879},
numpages = {17}
}

@ARTICLE{lr-survey-luevano-2019,
  author={Luevano, Luis S. and Chang, Leonardo and Méndez-Vázquez, Heydi and Martínez-Díaz, Yoanna and González-Mendoza, Miguel},
  journal={IEEE Access}, 
  title={A Study on the Performance of Unconstrained Very Low Resolution Face Recognition: Analyzing Current Trends and New Research Directions}, 
  year={2021},
  volume={9},
  number={},
  pages={75470-75493},
  doi={10.1109/ACCESS.2021.3080712}}

@article{guo2025gait,
  title={Gait recognition in the wild: A large-scale benchmark and nas-based baseline},
  author={Guo, Xianda and Zhu, Zheng and Yang, Tian and Lin, Beibei and Huang, Junjie and Deng, Jiankang and Huang, Guan and Zhou, Jie and Lu, Jiwen},
  journal={IEEE Transactions on Pattern Analysis and Machine Intelligence},
  year={2025},
  publisher={IEEE}
}

@inproceedings{gait3d,
title={Gait Recognition in the Wild with Dense 3D Representations and A Benchmark},
author={Jinkai Zheng and Xinchen Liu and Wu Liu and Lingxiao He and Chenggang Yan and Tao Mei},
booktitle={IEEE Conference on Computer Vision and Pattern Recognition (CVPR)},
year={2022}
}

@INPROCEEDINGS{uccs,
  author={G\"{u}nther, M. and Hu, P. and Herrmann, C. and Chan, C. H. and Jiang, M. and Yang, S. and Dhamija, A. R. and Ramanan, D. and Beyerer, J. and Kittler, J. and Jazaery, M. Al and Nouyed, M. I. and Guo, G. and Stankiewicz, C. and Boult, T. E.},
  booktitle={2017 IEEE International Joint Conference on Biometrics (IJCB)}, 
  title={Unconstrained Face Detection and Open-Set Face Recognition Challenge}, 
  year={2017},
  volume={},
  number={},
  pages={697-706},
  doi={10.1109/BTAS.2017.8272759}}

@article{martinez-diaz_benchmarking_fr_2021,
	title = {Benchmarking lightweight face architectures on specific face recognition scenarios},
	volume = {54},
	issn = {1573-7462},
	url = {https://doi.org/10.1007/s10462-021-09974-2},
	doi = {10.1007/s10462-021-09974-2},
	number = {8},
	journal = {Artificial Intelligence Review},
	author = {Martínez-Díaz, Yoanna and Nicolás-Díaz, Miguel and Méndez-Vázquez, Heydi and Luevano, Luis S. and Chang, Leonardo and Gonzalez-Mendoza, Miguel and Sucar, Luis Enrique},
	month = dec,
	year = {2021},
	pages = {6201--6244},
}

@article{survface,
title={Surveillance Face Recognition Challenge},
author={Cheng, Zhiyi and Zhu, Xiatian and Gong, Shaogang},
journal={arXiv preprint arXiv:1804.09691},
year={2018}
}

@article{george2024edgeface,
  title={Edgeface: Efficient face recognition model for edge devices},
  author={George, Anjith and Ecabert, Christophe and Shahreza, Hatef Otroshi and Kotwal, Ketan and Marcel, S{\'e}bastien},
  journal={IEEE Transactions on Biometrics, Behavior, and Identity Science},
  volume={6},
  number={2},
  pages={158--168},
  year={2024},
  publisher={IEEE}
}

@article{border_crossing_cbp_khan_2021,
title = {The use of biometric technology at airports: The case of customs and border protection (CBP)},
journal = {International Journal of Information Management Data Insights},
volume = {1},
number = {2},
pages = {100049},
year = {2021},
issn = {2667-0968},
doi = {https://doi.org/10.1016/j.jjimei.2021.100049},
url = {https://www.sciencedirect.com/science/article/pii/S2667096821000422},
author = {Nimra Khan and Marina Efthymiou}
}

@article{tinyface,
archivePrefix = {arXiv},
arxivId = {1811.08965},
author = {Cheng, Zhiyi and Zhu, Xiatian and Gong, Shaogang},
eprint = {1811.08965},
month = {nov},
pages = {1--16},
title = {{Low-Resolution Face Recognition}},
url = {http://arxiv.org/abs/1811.08965},
year = {2018},
journal={arXiv preprint arXiv:1811.08965}
}

@inproceedings{kim2022adaface,
  title={Adaface: Quality adaptive margin for face recognition},
  author={Kim, Minchul and Jain, Anil K and Liu, Xiaoming},
  booktitle={Proceedings of the IEEE/CVF conference on computer vision and pattern recognition},
  pages={18750--18759},
  year={2022}
}

@inproceedings{you2025lvface,
  title={LVFace: Progressive cluster optimization for large vision models in face recognition},
  author={You, Jinghan and Li, Shanglin and Sun, Yuanrui and Wei, Jiangchuan and Guo, Mingyu and Feng, Chao and Ran, Jiao},
  booktitle={Proceedings of the IEEE/CVF International Conference on Computer Vision},
  pages={11840--11849},
  year={2025}
}

@INPROCEEDINGS {lr_surveillance_martinez-diaz_2021,
author = { Martinez-Diaz, Yoanna and Mendez-Vazquez, Heydi and Luevano, Luis S. and Chang, Leonardo and Gonzalez-Mendoza, Miguel },
booktitle = { 2020 25th International Conference on Pattern Recognition (ICPR) },
title = {{ Lightweight Low-Resolution Face Recognition for Surveillance Applications }},
year = {2021},
volume = {},
ISSN = {1051-4651},
pages = {5421-5428},
doi = {10.1109/ICPR48806.2021.9412280},
url = {https://doi.ieeecomputersociety.org/10.1109/ICPR48806.2021.9412280},
publisher = {IEEE Computer Society},
address = {Los Alamitos, CA, USA},
month =Jan}

@misc{gdpr2016,
  title = {Regulation (EU) 2016/679 of the European Parliament and of the Council of 27 April 2016 on the protection of natural persons with regard to the processing of personal data and on the free movement of such data, and repealing Directive 95/46/EC (General Data Protection Regulation)},
  author = {{European Parliament} and {Council of the European Union}},
  year = {2016},
  journal = {Official Journal of the European Union},
  volume = {L 119},
  pages = {1--88},
  url = {http://data.europa.eu/eli/reg/2016/679/oj/eng}
}

@book{ai_act_2025,
author = {European Data Protection Supervisor},
title = {AI Act Regulation (EU) 2024/1689 - Regulation (EU) 2024/1689 of the European Parliament and of the Council of 13 June 2024 laying down harmonised rules on artificial intelligence and amending Regulations (EC) No 300/2008, (EU) No 167/2013, (EU) No 168/2013, (EU) 2018/858, (EU) 2018/1139 and (EU) 2019/2144 and Directives 2014/90/EU, (EU) 2016/797 and (EU) 2020/1828 (Artificial Intelligence Act) (Text with EEA relevance)},
publisher = {Publications Office of the European Union},
year = {2025},
doi = {doi/10.2804/4225375}}

@article{border_control_survey_labati_2016,
author = {Labati, Ruggero Donida and Genovese, Angelo and Mu\~{n}oz, Enrique and Piuri, Vincenzo and Scotti, Fabio and Sforza, Gianluca},
title = {Biometric Recognition in Automated Border Control: A Survey},
year = {2016},
issue_date = {June 2017},
publisher = {Association for Computing Machinery},
address = {New York, NY, USA},
volume = {49},
number = {2},
issn = {0360-0300},
url = {https://doi.org/10.1145/2933241},
doi = {10.1145/2933241},
journal = {ACM Comput. Surv.},
month = jun,
articleno = {24},
numpages = {39}
}

@article{forensic_gait_seckiner_2019,
title = {Forensic gait analysis — Morphometric assessment from surveillance footage},
journal = {Forensic Science International},
volume = {296},
pages = {57-66},
year = {2019},
issn = {0379-0738},
doi = {https://doi.org/10.1016/j.forsciint.2019.01.007},
url = {https://www.sciencedirect.com/science/article/pii/S0379073819300167},
author = {Dilan Seckiner and Xanthé Mallett and Philip Maynard and Didier Meuwly and Claude Roux}
}

@article{fitzpatrik_skin_gupta_2019,
title = {Skin typing: Fitzpatrick grading and others},
journal = {Clinics in Dermatology},
volume = {37},
number = {5},
pages = {430-436},
year = {2019},
note = {The Color of Skin},
issn = {0738-081X},
doi = {https://doi.org/10.1016/j.clindermatol.2019.07.010},
author = {Vishal Gupta and Vinod Kumar Sharma}
}

@inproceedings{droneface,
author = {Hsu, Hwai-Jung and Chen, Kuan-Ta},
title = {DroneFace: An Open Dataset for Drone Research},
year = {2017},
isbn = {9781450350020},
publisher = {Association for Computing Machinery},
address = {New York, NY, USA},
url = {https://doi.org/10.1145/3083187.3083214},
doi = {10.1145/3083187.3083214},
booktitle = {Proceedings of the 8th ACM on Multimedia Systems Conference},
pages = {187–192},
numpages = {6},
location = {Taipei, Taiwan},
series = {MMSys'17}
}

@article{fan2023exploring,
  title={Exploring deep models for practical gait recognition},
  author={Fan, Chao and Hou, Saihui and Huang, Yongzhen and Yu, Shiqi},
  journal={arXiv preprint arXiv:2303.03301},
  year={2023}
}

@misc{liu2025person,
      title={Person Recognition at Altitude and Range: Fusion of Face, Body Shape and Gait}, 
      author={Feng Liu and Nicholas Chimitt and Lanqing Guo and Jitesh Jain and Aditya Kane and Minchul Kim and Wes Robbins and Yiyang Su and Dingqiang Ye and Xingguang Zhang and Jie Zhu and Siddharth Satyakam and Christopher Perry and Stanley H. Chan and Arun Ross and Humphrey Shi and Zhangyang Wang and Anil Jain and Xiaoming Liu},
      year={2025},
      eprint={2505.04616},
      archivePrefix={arXiv},
      primaryClass={cs.CV},
      url={https://arxiv.org/abs/2505.04616}, 
}

@misc{wang2025combo,
      title={Combo-Gait: Unified Transformer Framework for Multi-Modal Gait Recognition and Attribute Analysis}, 
      author={Zhao-Yang Wang and Zhimin Shao and Anirudh Nanduri and Basudha Pal and Laura McDaniel and Jieneng Chen and Rama Chellappa},
      year={2026},
      eprint={2510.10417},
      archivePrefix={arXiv},
      primaryClass={cs.CV},
      url={https://arxiv.org/abs/2510.10417}, 
}

@inproceedings{geng2008adaptive,
author = {Geng, Xin and Wang, Liang and Li, Ming and Wu, Qiang and Smith-Miles, Kate},
title = {Adaptive Fusion of Gait and Face for Human Identification in Video},
year = {2008},
isbn = {9781424419135},
publisher = {IEEE Computer Society},
address = {USA},
url = {https://doi.org/10.1109/WACV.2008.4544006},
doi = {10.1109/WACV.2008.4544006},
booktitle = {Proceedings of the 2008 IEEE Workshop on Applications of Computer Vision},
pages = {1–6},
numpages = {6},
series = {WACV '08}
}

@inproceedings{fu2022fusion,
author = {Fu, Hui and Kang, Wenxiong and Zhang, Yuxuan and Shakeel, M. Saad},
title = {Fusion of Gait and Face for Human Identification at the Feature Level},
year = {2022},
isbn = {978-3-031-20232-2},
publisher = {Springer-Verlag},
address = {Berlin, Heidelberg},
url = {https://doi.org/10.1007/978-3-031-20233-9_48},
doi = {10.1007/978-3-031-20233-9_48},
booktitle = {Biometric Recognition: 16th Chinese Conference, CCBR 2022, Beijing, China, November 11–13, 2022, Proceedings},
pages = {475–483},
numpages = {9},
location = {Beijing, China}
}

@INPROCEEDINGS{muramatsu2013multi,
  author={Muramatsu, Daigo and Iwama, Haruyuki and Makihara, Yasushi and Yagi, Yasushi},
  booktitle={2013 International Conference on Biometrics (ICB)}, 
  title={Multi-view multi-modal person authentication from a single walking image sequence}, 
  year={2013},
  volume={},
  number={},
  pages={1-8},
  doi={10.1109/ICB.2013.6612979}
  }

@INPROCEEDINGS{kale2004fusion,
  author={Kale, A. and Roychowdhury, A.K. and Chellappa, R.},
  booktitle={2004 IEEE International Conference on Acoustics, Speech, and Signal Processing}, 
  title={Fusion of gait and face for human identification}, 
  year={2004},
  volume={5},
  number={},
  pages={V-901},
  doi={10.1109/ICASSP.2004.1327257}}

@Article{maity2021multimodal,
AUTHOR = {Maity, Sayan and Abdel-Mottaleb, Mohamed and Asfour, Shihab S.},
TITLE = {Multimodal Low Resolution Face and Frontal Gait Recognition from Surveillance Video},
JOURNAL = {Electronics},
VOLUME = {10},
YEAR = {2021},
NUMBER = {9},
ARTICLE-NUMBER = {1013},
URL = {https://www.mdpi.com/2079-9292/10/9/1013},
ISSN = {2079-9292},
DOI = {10.3390/electronics10091013}
}

@Article{aung2022multimodal,
AUTHOR = {Aung, Hsu Mon Lei and Pluempitiwiriyawej, Charnchai and Hamamoto, Kazuhiko and Wangsiripitak, Somkiat},
TITLE = {Multimodal Biometrics Recognition Using a Deep Convolutional Neural Network with Transfer Learning in Surveillance Videos},
JOURNAL = {Computation},
VOLUME = {10},
YEAR = {2022},
NUMBER = {7},
ARTICLE-NUMBER = {127},
URL = {https://www.mdpi.com/2079-3197/10/7/127},
ISSN = {2079-3197},
DOI = {10.3390/computation10070127}
}
